\documentclass[11pt]{article}
\newcommand{\ci}[2]{\,{\fontsize{5pt}{7.2pt}\selectfont[#1,#2]}} 
\usepackage[final]{acl}

\usepackage{times}
\usepackage{latexsym}
\usepackage{multirow}
\usepackage{array}
\usepackage{tcolorbox}
\usepackage{subcaption}
\usepackage{booktabs, multirow, makecell, colortbl, xcolor, adjustbox, array}

\usepackage[T2A]{fontenc}
\usepackage[T1]{fontenc}

\usepackage{amsmath}
\usepackage{fontspec}
\usepackage{amssymb}
\usepackage{adjustbox}
\usepackage{float}
\newfontfamily\kalpurush{kalpurush.ttf}
\usepackage[english,bidi=default]{babel} 
\babelfont{rm}{TeXGyreTermesX} 
\babelprovide[import]{hindi}
\babelprovide[import]{arabic}
\babelfont[*arabic]{rm}{Noto Sans Arabic}
\babelprovide[import]{hindi}
\newcommand{\banglafont}[1]{{\kalpurush #1}}
\usepackage{xcolor}
\definecolor{bottlegreen}{RGB}{0,106,78}

\usepackage[normalem]{ulem}   
\newif\ifshowrev
\showrevfalse  
\ifshowrev
  \DeclareRobustCommand{\rev}[1]{\leavevmode{\color{red}{}#1}}
  \DeclareRobustCommand{\del}[1]{\leavevmode{\color{red}\sout{#1}}}
\else
  \DeclareRobustCommand{\rev}[1]{#1}
  \DeclareRobustCommand{\del}[1]{\ignorespaces}
\fi

\usepackage{microtype}

\usepackage{inconsolata}

\usepackage{graphicx}
\usepackage{booktabs}

\title{\textsc{BanglaWild}: An In-the-Wild Bengali Scene Text Recognition Benchmark for OCR and Vision-Language Models}

\author{
Sadab Shiper$^{1}$\thanks{Equal contribution},
Tawsif Tashwar Dipto$^{1}$\footnotemark[1],
Mir Md Inzamam$^{1}$,
Eshat Tanzeem$^{2}$ \\
$^{1}$Islamic University of Technology, Bangladesh \\
$^{2}$BRAC University, Bangladesh \\
\texttt{sadabshiper@iut-dhaka.edu},
\texttt{tawsiftashwar@iut-dhaka.edu},\\
\texttt{mdinzamam@iut-dhaka.edu},
\texttt{eshat.tanzeem@g.bracu.ac.bd}
}

\begin{document}
\maketitle


\begin{abstract}
In-the-wild Bengali scene text recognition is largely unmeasured: existing
resources target handwritten documents or constrained signboard parsing,
report only aggregate edit-distance metrics, and evaluate either conventional
OCR or VLMs, never both on the same
in-the-wild data. We introduce \textsc{BanglaWild}, a
benchmark of 2{,}535 Bengali scene text images, each
paired with a verbatim gold transcription, two categorical axes, four
diagnostic attributes, and an orthographically standard form where the
in-image text deviates from canonical spelling. We evaluate fifteen VLMs and three conventional OCR systems under three
prompting strategies, fine-tune 6 open-source models with LoRA, and
complement edit-distance metrics with an LLM-as-a-Judge evaluation. Our results
reveal a persistent gap in which larger models within the same family
do not consistently outperform smaller ones. Our fifteen-class error taxonomy shows
that \del{visual mis-recognition accounts for \textasciitilde{}60\% of errors} \rev{$\sim$60\% of word-level errors} in the strongest
systems\rev{ are consistent with visual mis-recognition}, while conjunct-related errors contribute under 2\%, challenging a
long-standing assumption in Bengali OCR research; the same visual-dominant
profile \del{also} holds \del{across architectures, including} \rev{for EasyOCR,} the one conventional baseline that reads Bengali reliably\rev{, before and after fine-tuning}. Prompt language
mainly affects cross-script drift and LoRA reduces catastrophic failures in
weak models without lifting the ceiling on already competent ones. Code and data \del{will be publicly released} \rev{are available at} \url{https://github.com/SadabShiper/BanglaWild}.
\end{abstract}

\begin{figure*}[t]
    \centering
    \includegraphics[width=1.0\textwidth, page=1]{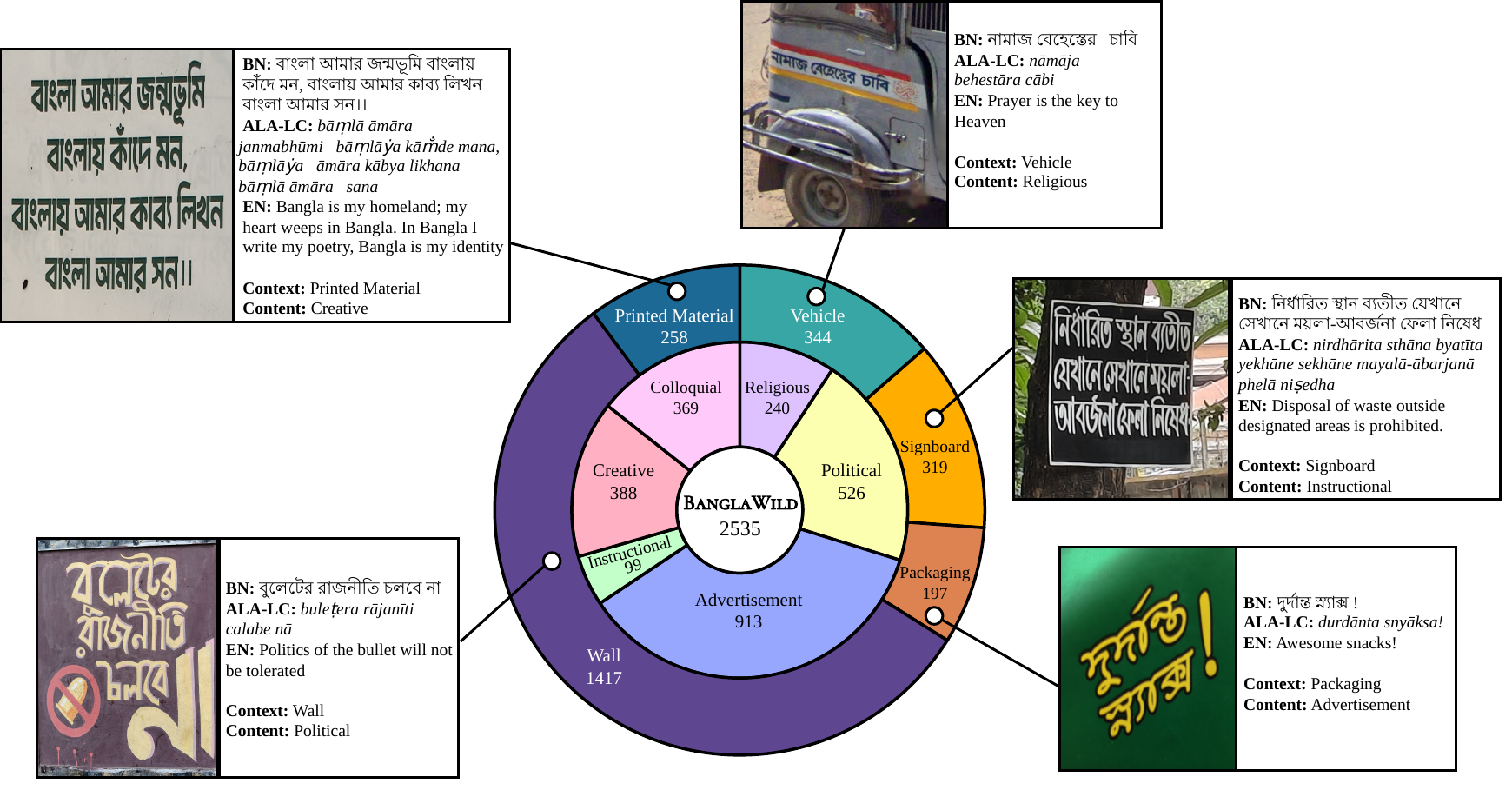}
    \caption{Category-wise distribution of \textsc{BanglaWild} across \textsc{Context} (outer 
ring) and \textsc{Content} (inner ring) axes, with representative image-annotation 
pairs for each content category. Each annotation includes the original Bengali text (BN), an ALA-LC Romanization \citep{alalc2017bengali} added here 
for readability, and an English translation (EN), neither of which is part of the released dataset.}
    \label{fig:chakti}
\vspace{-3mm}

\end{figure*}

\vspace{-3mm}
\section{Introduction}
\label{sec:intro}
Vision-language models (VLMs) \citep{gemini2_5_2025, qwen3_5_2026} are advertised as broadly multilingual, yet the evidence supporting that claim skews sharply toward English and Chinese text, toward document OCR rather than in-the-wild scene text, and toward short, clean image-text pairs rather than the cluttered surfaces on which text actually appears in public space \citep{fu2024ocrbenchv2improvedbenchmark, tang2024mtvqa}. For low-resource Indic scripts, the gap between marketing and measurement is particularly acute. Although VLMs are routinely deployed in settings where Bengali scene text matters (transcribing addresses, receipts, election material, and medical packaging), we have no rigorous account of how, or how badly, they fail. Existing Bengali resources target conventional OCR settings such as cropped-word recognition or handwritten documents under controlled conditions \citep{rahman2023bnhtrd}, leaving the perspective distortion, stylized typography, multilingual clutter, and occlusion that characterize Bengali text on public walls, street vehicles, banners, and product packaging entirely unmeasured. Crucially, no existing Bengali resource evaluates conventional OCR and generative VLMs on the same in-the-wild data, so it remains unknown whether modern VLMs improve over conventional OCR on this setting at all.
We introduce \textsc{BanglaWild}, a diagnostic benchmark of 2{,}535 natively photographed Bengali scene text images captured on public-facing physical surfaces under uncontrolled conditions. Each image is paired with a verbatim gold transcription that preserves any in-image spelling errors or non-standard glyph substitutions, two categorical labels (\textsc{Context}, \textsc{Content}), and four diagnostic attributes (\textsc{Curved Text}, \textsc{Font Style}, \textsc{Occlusion}, \textsc{Background Complexity}). For the 3.43\% of images whose in-image text deviates from standard spelling, an orthographically standard form is recorded in a separate field. This dual annotation lets us distinguish \emph{recognition} errors (the model misreads what is present) from \emph{correction} errors (the model silently rewrites a misspelling into its standard form), a distinction that no prior Bengali scene text benchmark supports.
Our contributions are as follows:
\begin{itemize}
    \vspace{-2mm}
    \item We present the first in-the-wild Bengali scene text recognition benchmark evaluating both conventional OCR and generative VLMs, stratified by categorical and diagnostic attributes with adjudicated gold transcriptions. Although the surface task is verbatim transcription, its dual verbatim/standard annotation targets two failure modes conventional OCR cannot structurally exhibit: silent over-correction, since a CRNN has no lexical prior to normalize a misspelling toward, and fluent hallucination toward plausible Bengali that a reader cannot flag from the output alone. We evaluate OCR alongside VLMs not because the task reduces to OCR, but to \del{show} \rev{test whether} the dominant failure mode is \del{a property of the imagery, not} \rev{specific to} one model class.
    \vspace{-2mm}
    \item We benchmark 15 contemporary VLMs and 3 conventional OCR systems across 3 prompting strategies \del{ and}\rev{,} fine-tune 6 open-source models with LoRA\rev{, and adapt EasyOCR's recognizer on the same training split}, complemented by a validated LLM-as-a-Judge protocol on the full test set.
    \vspace{-2mm}
    \item We provide a fifteen-class error taxonomy that identifies \rev{errors consistent with} visual mis-recognition, not orthographic confusion, as \del{the dominant failure mode in Bengali scene text recognition, which holds across VLMs and a conventional OCR baseline, and bounds what prompt engineering and fine-tuning can achieve} \rev{dominant among word-level errors in four VLM families under all three prompts and in EasyOCR before and after fine-tuning}.
\end{itemize}

\begin{figure*}[t]
    \centering
    \includegraphics[width=0.8\textwidth,page=1]{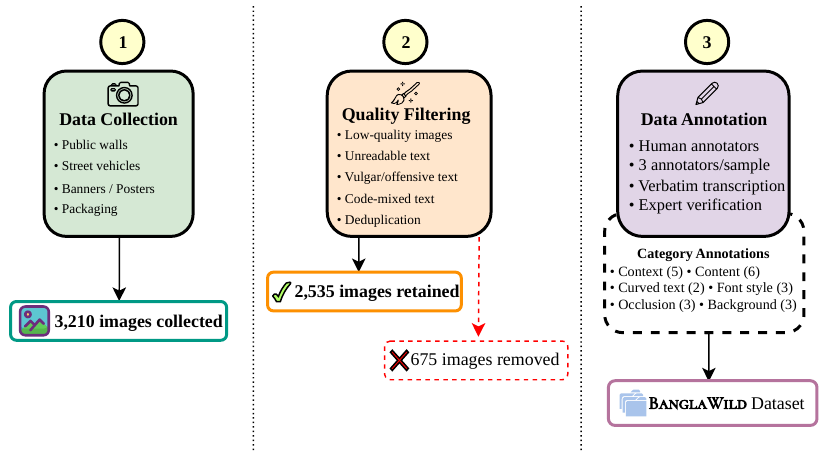}
    \caption{Overview of the \textsc{BanglaWild} Data Collection and Annotation Pipeline.}
    \label{fig:pipeline}
    \vspace{-4mm}
\end{figure*}

\vspace{-3mm}
\section{Related Work}
\label{sec:related}
\vspace{-3mm}
\paragraph{Indic and Bengali Scene Text Recognition.}
    Multilingual scene text understanding for Indic scripts has been driven by the ICDAR Robust Reading competitions RRC-MLT-2017 and RRC-MLT-2019 \citep{nayef2019icdar}, which standardized multi-script benchmarks and evaluation protocols. Building on these, \citet{gunna2022transferlearning} investigate cross-script transfer learning from English to Hindi and Gujarati with CRNN and STAR-Net, evaluating transfer among six Indic languages, and show that cross-Indic transfer outperforms English-to-Indic transfer. For Bengali, public scene text resources remain scarce and target conventional OCR settings, including crowdsourced collection \citep{hossain2023crowdsource} and structured signboard parsing \citep{murad2023bangla}. Other Bengali resources target handwritten document recognition rather than scene text, including BN-HTRd \citep{rahman2023bnhtrd}, the largest document-level Bengali HTR corpus with 108{,}147 annotated words across 788 pages, which remains the canonical reference for handwritten Bengali but is captured under controlled writing conditions. More recently, \citet{de2025bharat} introduced BSTD, the largest multilingual Indic scene text benchmark to date with 6{,}582 images and 126{,}292 annotated words across 11 languages; only 6{,}304 ($\sim$5\%) are Bengali, sourced from Wikimedia Commons rather than captured in-the-wild. These resources are built for CRNN or encoder-decoder pipelines rather than generative VLMs, report only aggregate CER and WER, and do not isolate the script drift, conjunct degradation, or cross-script interference that uniquely affect generative VLMs on Indic scripts. We therefore do not claim the first Bengali scene text or OCR resource: \citet{hossain2023crowdsource} introduced a crowdsourced detection pipeline, and \citet{murad2023bangla} proposed a structured signboard address parser, both valuable prior work. \textsc{BanglaWild} is instead the first in-the-wild Bengali scene text recognition benchmark to pair verbatim/standard dual annotation with evaluation of both conventional OCR and generative VLMs on the same data, none of which these resources provide.
\vspace{-2mm}
\paragraph{Text-Centric and Indic VLM Benchmarks.}
Text-centric VLM benchmarks have expanded evaluation beyond classical OCR, with OCRBench \citep{Liu_2024} first systematically evaluating multimodal models on OCR-centric tasks. OCRBench~v2 \citep{fu2024ocrbenchv2improvedbenchmark} extended this with 10{,}000 human-verified QA pairs across 31 scenarios, exposing persistent weaknesses in localization and fine-grained perception. Both versions are English and Chinese focused (OCRBench v2 covers 31 scenarios but no Indic-script content) and thus do not surface the script-level failure modes relevant to Bengali. MTVQA \citep{tang2024mtvqa} introduced the first multilingual text-centric VQA benchmark and identified the visual-textual misalignment problem from translated rather than visually grounded text, a constraint we adopt by sourcing only natively photographed Bengali scenes; MTVQA itself excludes Bengali and other Indic scripts. ThaiOCRBench \citep{nonesung2025thaiocrbench} showed that low-resource scripts expose substantial hallucination and robustness failures in modern VLMs, directly motivating our cross-script analysis. \citet{sohail2024lowresource} report findings on 2{,}520 controlled images of Urdu, Albanian, and Tajik, showing that script complexity, not low-resource status alone, drives GPT-4o \cite{OpenAI2024GPT4o} failure. Concurrent work has identified the related failure mode of VLM \emph{over-correction} in handwritten transcription \citep{seong2026vlmsfixstudentsidentifying}, where models silently fix errors rather than transcribing them verbatim; our dual annotation extends this diagnostic capability to Bengali scene text, where orthographic non-standardness is the norm rather than the exception. Most relevant to our setting, IndicVisionBench \citep{faraz2025indicvisionbench} provides the first large-scale Indic VLM benchmark, covering OCR, MMT, and VQA over English and 10 Indian languages through $\sim$5{,}000 culturally grounded images and 37{,}000+ QA pairs. Its per-language OCR allocation is necessarily small once split across 11 languages and three tracks, and its imagery centers on cultural topics rather than uncontrolled urban scenes, so the benchmark emphasizes cultural understanding and cross-lingual evaluation over fine-grained script-level failure analysis on in-the-wild Bengali text.

\section{The \textsc{BanglaWild} Dataset}
\label{sec:dataset}

We introduce \textsc{BanglaWild}, a benchmark of \textbf{2{,}535} in-the-wild images of Bengali scene text captured on public-facing physical surfaces under uncontrolled conditions. Each image includes a verbatim gold transcription, two categorical labels (\textsc{Context}, \textsc{Content}), four diagnostic attributes (\textsc{Curved Text}, \textsc{Font Style}, \textsc{Occlusion}, \textsc{Background Complexity}), and an orthographically standard form when the in-image text deviates from standard spelling. Details of the diagnostic attributes are provided in Appendix~\ref{app:rubric}. Fig.~\ref{fig:pipeline} summarizes the construction pipeline, and Fig.~\ref{fig:chakti} reports category-wise statistics.

\paragraph{Image Sourcing.}
Photographs were captured by trained collaborators using consumer mobile-phone cameras across five sources\rev{, chosen as the most common carriers of public Bengali text in the regions of Bangladesh we sampled}: \textit{public walls}, \textit{signboards}, \textit{street vehicles}, \textit{banners and posters}, and \textit{consumer product packaging}. \rev{To mirror deployment conditions, we did not standardize devices, lighting, or capture settings, so images retain their natural lighting and naturally occurring text quality.} From an initial pool of \textbf{3{,}210} images, we \rev{and our collaborators} discarded \textbf{675} ($21.0\%$) that were (i) too low-quality or weathered for confident transcription, (ii) unreadable despite acceptable image quality, (iii) dominated by code-mixed Bengali-English text\rev{, which is outside our scope}, or (iv) vulgar or offensive, yielding the final \textbf{2,535}-image benchmark. \rev{Criteria (i) and (ii) were judged by whether our annotators could transcribe an image confidently, since otherwise no reliable gold reference exists; because criteria (i)--(iii) exclude some of the hardest real-world inputs, our error rates likely understate the difficulty of unfiltered deployment data (see Limitations).} Filtering also included deduplication: near-identical images sharing both text and visual appearance \rev{(writing style, surface, lighting, and image quality)} were merged into a single representative, whereas identical text with different surfaces, fonts, layouts, or lighting was retained because it poses distinct recognition challenges. Incidentally captured faces and vehicle registration plates were blurred and verified as non-identifying.

\paragraph{Data Annotation.}
Each image is labeled along two orthogonal categorical axes. \textsc{Context} describes the physical medium on which text appears (\textit{Vehicle}, \textit{Signboard}, \textit{Packaging}, \textit{Wall}, \textit{Printed Material}), while \textsc{Content} captures its communicative intent (\textit{Religious}, \textit{Political}, \textit{Advertisement}, \textit{Instructional}, \textit{Creative}, \textit{Colloquial}). Formal definitions are provided in Appendix~\ref{app:categories}.

We recruited \textbf{three} native Bengali-speaking annotators (university undergraduates with Bengali literature backgrounds, compensated at standard per-sample rates) to independently transcribe every image. Following the rubric (Appendix~\ref{app:rubric}), the gold transcription preserves the text \emph{exactly as it appears}, including spelling errors and non-standard glyph substitutions. A separate field records the orthographically standard form for the \textbf{87} images ($3.43\%$) containing non-standard spellings. This dual annotation distinguishes \emph{recognition errors} (misreading the image) from \emph{correction errors} (silently normalizing misspellings). For each of the 15 models under all 3 prompt variants, we classify outputs as matching the verbatim transcription, the standardized form, or neither (Tab.~\ref{tab:exact-match-perprompt}). Separately, two annotators labeled the \textsc{Context}, \textsc{Content}, and four diagnostic attributes, with disagreements resolved by an expert in Bengali orthography. We use a 50/10/40 train/validation/test split (1,268/253/1,014 samples), evaluate all models on the common test split, and report results stratified across all six label dimensions in Tabs.~\ref{tab:bangla_wild_results} and~\ref{tab:bangla_wild_attributes}.

\paragraph{Annotation Verification.}
Disagreements among the three transcribers were resolved by the expert verifier to produce the canonical gold transcription. Overall, $18.4\%$ of images required adjudication on at least one character; the remainder achieved unanimous first-pass agreement. Mean pairwise CER across all 2{,}535 images is $3.2\%$, and Cohen's $\kappa$ ranges from $0.81$--$0.93$ across all categorical labels, indicating strong agreement (full breakdown in Tab.~\ref{tab:iaa}).

\section{Experiment Design}
\label{sec:eval}
We benchmark 15 vision-language models (full list in Tab.~\ref{tab:models}) under two regimes, (i) zero-shot prompting on the full \textsc{BanglaWild} test set and (ii) LoRA fine-tuning of \del{the} 6 open-source models with publicly available weights, and additionally evaluate 3 conventional OCR systems as non-generative baselines, scored with the same pipeline as the VLM outputs. \rev{To check whether the OCR error profile merely reflects missing domain adaptation, we also fine-tune EasyOCR's recognizer on the same training split (Appendix~\ref{app:training}).} We report CER, WER, and $1-\text{NED}$ in Tab.~\ref{tab:leaderboard}, and additionally an LLM-as-a-Judge score \citep{zheng2023judging} on a 1-5 rubric-anchored scale to complement edit-distance metrics with a meaning-aware signal; the judge prompt and validation protocol are in Appendix~\ref{app:judge_prompt}.
\subsection{Zero-Shot Prompting}
\label{sec:zero_shot}
We evaluate using three prompting strategies: English and Bengali baseline prompts, and a structured metadata prompt. \textsc{P1} and \textsc{P2} are image-only conditions and constitute our deployment-realistic setting. \textsc{P3} additionally prepends the gold \textsc{Context} and \textsc{Content} labels, which are annotator-provided and unavailable at inference time; we therefore treat \textsc{P3} as an \emph{oracle condition} that upper-bounds the value of perfect scene priors, report it for reference only, and base all headline claims and best-prompt comparisons on \{\textsc{P1}, \textsc{P2}\}. Prompting details and examples are in Appendix~\ref{app:prompts}.
\subsection{Parameter-Efficient Fine-Tuning}
\label{sec:lora}
We apply LoRA~\citep{hu2022lora} to test whether targeted adaptation improves Bengali scene text recognition beyond zero-shot prompting. For a pretrained weight matrix $\mathbf{W}_0 \in \mathbb{R}^{d \times k}$, LoRA freezes $\mathbf{W}_0$ and learns a low-rank update
\[
\mathbf{W}_0 + \Delta\mathbf{W} \;=\; \mathbf{W}_0 + \mathbf{B}\mathbf{A},
\]
where $\mathbf{B} \in \mathbb{R}^{d \times r}$, $\mathbf{A} \in \mathbb{R}^{r \times k}$, and $r \ll \min(d, k)$. Only $\mathbf{A}$ and $\mathbf{B}$ receive gradient updates. At inference we use the merged weights $\mathbf{W} = \mathbf{W}_0 + \mathbf{B}\mathbf{A}$, incurring no additional latency. Each model is trained on (image, prompt) pairs against the ground-truth transcription using only \textsc{P1}. LoRA adapters are applied to the attention projections ($q\_proj$, $k\_proj$, $v\_proj$, $o\_proj$) and MLP projections ($up\_proj$, $down\_proj$, $gate\_proj$) of the \emph{language-model component only}; the vision encoder and any vision-language connector are kept frozen throughout training. Training settings are detailed in Appendix~\ref{app:training}.

\section{Result Analysis}
\label{sec:results}

\begin{table*}[h]
\scriptsize
\centering
\setlength\tabcolsep{2pt}
\renewcommand{\arraystretch}{1.12}
\resizebox{2.1\columnwidth}{!}{
\begin{tabular}{@{}l ccc ccc ccc ccc@{}}
\toprule
& \multicolumn{3}{c}{\textbf{CER}\textcolor{green}{$\downarrow$}}
& \multicolumn{3}{c}{\textbf{WER}\textcolor{green}{$\downarrow$}}
& \multicolumn{3}{c}{\textbf{1$-$NED}\textcolor{green}{$\uparrow$}} 
& \multicolumn{3}{c}{\textbf{Judge}\textcolor{green}{$\uparrow$}} \\
\cmidrule(lr){2-4} \cmidrule(lr){5-7} \cmidrule(lr){8-10} \cmidrule(l){11-13}
\textbf{Model} & P1 & P2 & P3$^{\dagger}$ & P1 & P2 & P3$^{\dagger}$ & P1 & P2 & P3$^{\dagger}$ & P1 & P2 & P3$^{\dagger}$ \\
\midrule

\multicolumn{13}{c}{\textbf{\textit{Zero-Shot Prompting}}} \\
\cmidrule(lr){1-13}

\multicolumn{13}{@{}l}{\textbf{\textit{Closed-source VLMs}}} \\

Gemini 2.5 Flash
& \textbf{14.08}\ci{12.62}{15.54}
& 15.60\ci{13.69}{17.51}
& 13.93\ci{12.63}{15.23}
& \textbf{29.97}\ci{28.08}{31.86}
& 31.37\ci{29.11}{33.63}
& 30.22\ci{28.39}{32.05}
& 82.11\ci{81.13}{83.09}
& \textbf{82.55}\ci{81.61}{83.49}
& 82.25\ci{81.28}{83.22}
& $\text{3.68}^{*}$
& $\textbf{3.69}^{*}$
& $\text{3.70}^{*}$ \\

Claude Sonnet 4.6
& 23.15\ci{17.02}{29.28}
& \textbf{18.46}\ci{13.12}{23.80}
& 17.96\ci{14.47}{21.45}
& 43.58\ci{37.77}{49.39}
& \textbf{39.67}\ci{34.89}{44.45}
& 38.50\ci{35.18}{41.82}
& 78.89\ci{77.81}{79.97}
& \textbf{80.12}\ci{79.10}{81.14}
& 79.47\ci{78.40}{80.54}
& 3.78
& \textbf{3.82}
& 3.81 \\

Gemini 2.5 Pro
& 24.85\ci{23.26}{26.44}
& \textbf{21.65}\ci{20.11}{23.19}
& 23.95\ci{22.31}{25.59}
& 39.53\ci{37.50}{41.56}
& \textbf{36.42}\ci{34.52}{38.32}
& 38.29\ci{36.32}{40.26}
& 76.10\ci{74.87}{77.33}
& \textbf{78.32}\ci{77.12}{79.52}
& 76.94\ci{75.69}{78.19}
& $\text{3.58}^{*}$
& $\textbf{3.67}^{*}$
& $\text{3.63}^{*}$ \\

Llama 4 Maverick
& \textbf{36.96}\ci{33.20}{40.72}
& 37.98\ci{34.20}{41.76}
& 37.74\ci{33.9}{41.6}
& \textbf{59.43}\ci{53.19}{65.67}
& 60.60\ci{54.2}{67.1}
& 59.89\ci{52.95}{66.83}
& 58.69\ci{57.33}{60.05}
& \textbf{60.61}\ci{59.34}{61.88}
& 58.73\ci{57.37}{60.09}
& 2.65
& \textbf{2.66}
& 2.65 \\

GPT-5.4-mini
& \textbf{53.41}\ci{50.84}{55.98}
& 56.83\ci{52.54}{61.12}
& 52.49\ci{49.65}{55.33}
& \textbf{81.17}\ci{78.90}{83.44}
& 85.14\ci{80.96}{89.32}
& 80.11\ci{77.22}{83.00}
& 42.70\ci{41.45}{43.95}
& \textbf{44.19}\ci{43.04}{45.34}
& 43.03\ci{41.81}{44.25}
& \textbf{1.77}
& \textbf{1.77}
& 1.80 \\

Claude Haiku 4.5
& 90.47\ci{66.17}{114.77}
& \textbf{82.17}\ci{66.89}{97.45}
& 88.72\ci{70.40}{107.04}
& 113.33\ci{94.03}{132.63}
& \textbf{108.68}\ci{93.26}{124.10}
& 113.45\ci{95.53}{131.37}
& 28.44\ci{27.36}{29.52}
& \textbf{32.09}\ci{31.12}{33.06}
& 28.45\ci{27.36}{29.54}
& 2.15
& \textbf{2.25}
& 2.22 \\

GPT-5.4-nano
& 246.95\ci{218.4}{280.7}
& \textbf{192.80}\ci{170.2}{220.8}
& 114.82\ci{105.5}{125.1}
& 349.15\ci{303.74}{394.56}
& \textbf{236.03}\ci{205.88}{266.18}
& 140.50\ci{131.18}{149.82}
& 11.53\ci{10.94}{12.12}
& \textbf{15.25}\ci{14.70}{15.80}
& 11.43\ci{10.96}{11.90}
& 1.02
& \textbf{1.08}
& 1.09 \\

\multicolumn{13}{@{}l}{\textbf{\textit{Open-source VLMs}}} \\

Qwen3.5-9B
& \textbf{20.38}\ci{19.97}{20.79}
& 20.51\ci{17.36}{23.66}
& 20.04\ci{18.65}{21.43}
& \textbf{45.23}\ci{43.52}{46.94}
& 45.90\ci{43.12}{48.68}
& 44.91\ci{43.10}{46.72}
& 74.68\ci{73.67}{75.69}
& \textbf{75.57}\ci{74.61}{76.53}
& 74.83\ci{73.82}{75.84}
& 3.29
& \textbf{3.30}
& \rev{3.30} \\

Qwen3.5-35B-A3B
& 21.20\ci{20.89}{21.51}
& \textbf{20.23}\ci{18.04}{22.42}
& 20.45\ci{19.02}{21.88}
& 43.94\ci{42.22}{45.66}
& \textbf{42.87}\ci{40.04}{45.70}
& 43.01\ci{41.14}{44.88}
& 72.29\ci{71.15}{73.43}
& \textbf{74.23}\ci{73.19}{75.27}
& 73.21\ci{72.08}{74.34}
& 3.18
& \textbf{3.22}
& 3.20 \\

Qwen3.5-27B
& 23.87\ci{22.23}{25.51}
& \textbf{23.45}\ci{21.90}{25.00}
& 23.71\ci{22.11}{25.31}
& 48.62\ci{44.50}{52.74}
& \textbf{47.66}\ci{45.78}{49.54}
& 47.77\ci{45.89}{49.65}
& 70.29\ci{69.15}{71.43}
& \textbf{71.55}\ci{70.50}{72.60}
& 70.66\ci{69.53}{71.79}
& 3.08
& \textbf{3.12}
& 3.10 \\

Gemma-3-12B-it
& \textbf{63.89}\ci{49.44}{78.34}
& 70.76\ci{48.44}{93.08}
& 52.35\ci{47.16}{57.54}
& \textbf{81.90}\ci{70.19}{93.61}
& 89.24\ci{74.14}{104.34}
& 74.34\ci{71.41}{77.27}
& 50.47\ci{49.32}{51.62}
& \textbf{52.39}\ci{51.28}{53.50}
& 50.84\ci{49.66}{52.02}
& 2.00
& \textbf{2.03}
& 2.04 \\

Gemma-3-4B-it
& 134.82\ci{105.35}{164.29}
& \textbf{119.64}\ci{75.42}{163.86}
& 72.38\ci{66.21}{78.55}
& \textbf{110.31}\ci{95.60}{125.02}
& 121.18\ci{95.92}{146.44}
& 94.34\ci{87.92}{100.76}
& 39.96\ci{38.90}{41.02}
& \textbf{42.50}\ci{41.48}{43.52}
& 40.34\ci{39.33}{41.35}
& 1.48
& \textbf{1.55}
& 1.62 \\

InternVL3.5-8B
& \textbf{116.37}\ci{103.99}{128.75}
& 116.87\ci{105.2}{128.5}
& 104.01\ci{93.5}{114.5}
& \textbf{127.46}\ci{108.88}{146.04}
& 131.70\ci{113.5}{136.70}
& 118.61\ci{112.98}{124.24}
& 12.97\ci{12.61}{13.33}
& \textbf{17.30}\ci{16.98}{17.62}
& 11.71\ci{11.35}{12.07}
& \textbf{1.01}
& \textbf{1.01}
& 1.00 \\

InternVL3.5-4B
& \textbf{118.44}\ci{111.20}{125.68}
& 156.92\ci{148.19}{165.65}
& 115.40\ci{105.0}{125.8}
& \textbf{134.02}\ci{125.27}{142.77}
& 172.92\ci{163.73}{182.11}
& 137.74\ci{133.19}{142.29}
& \textbf{17.32}\ci{17.00}{17.64}
& 16.43\ci{16.12}{16.74}
& 7.27\ci{7.04}{7.50}
& \textbf{1.02}
& 1.00
& 1.00 \\

InternVL3.5-2B
& \textbf{132.73}\ci{125.69}{139.77}
& 176.70\ci{165.69}{187.71}
& 136.05\ci{128.00}{144.10}
& \textbf{153.86}\ci{146.29}{161.43}
& 202.01\ci{191.93}{212.09}
& 151.79\ci{144.50}{159.08}
& \textbf{16.36}\ci{16.04}{16.68}
& 15.57\ci{15.25}{15.89}
& 11.69\ci{11.35}{12.03}
& \textbf{1.01}
& 1.00
& 1.00 \\

\midrule
\multicolumn{13}{c}{\textbf{\textit{LoRA Fine-Tuning (P1 Prompt)}}} \\
\cmidrule(lr){1-13}
& \multicolumn{3}{c}{\textbf{CER}\textcolor{green}{$\downarrow$}}
& \multicolumn{3}{c}{\textbf{WER}\textcolor{green}{$\downarrow$}}
& \multicolumn{3}{c}{\textbf{1$-$NED}\textcolor{green}{$\uparrow$}}
& \multicolumn{3}{c}{\textbf{Judge}\textcolor{green}{$\uparrow$}} \\
\cmidrule(lr){2-4} \cmidrule(lr){5-7} \cmidrule(lr){8-10} \cmidrule(l){11-13}

Qwen3.5-9B (FT)
& \multicolumn{3}{c}{21.53\ci{19.82}{23.24}}
& \multicolumn{3}{c}{40.16\ci{38.41}{41.91}}
& \multicolumn{3}{c}{75.93\ci{74.98}{76.88}}
& \multicolumn{3}{c}{3.25} \\

Gemma-3-12B-it (FT)
& \multicolumn{3}{c}{28.91\ci{25.84}{31.98}}
& \multicolumn{3}{c}{44.29\ci{40.88}{47.70}}
& \multicolumn{3}{c}{72.79\ci{71.44}{74.14}}
& \multicolumn{3}{c}{2.85} \\

InternVL3.5-8B (FT)
& \multicolumn{3}{c}{45.81\ci{41.37}{50.25}}
& \multicolumn{3}{c}{74.60\ci{69.28}{79.92}}
& \multicolumn{3}{c}{59.07\ci{57.24}{60.90}}
& \multicolumn{3}{c}{1.85} \\

Gemma-3-4B-it (FT)
& \multicolumn{3}{c}{50.72\ci{45.11}{56.33}}
& \multicolumn{3}{c}{65.82\ci{60.14}{71.50}}
& \multicolumn{3}{c}{69.50\ci{67.88}{71.12}}
& \multicolumn{3}{c}{1.70} \\

InternVL3.5-4B (FT)
& \multicolumn{3}{c}{51.73\ci{46.82}{56.64}}
& \multicolumn{3}{c}{77.42\ci{71.33}{83.51}}
& \multicolumn{3}{c}{51.53\ci{49.61}{53.45}}
& \multicolumn{3}{c}{1.50} \\

InternVL3.5-2B (FT)
& \multicolumn{3}{c}{51.96\ci{47.88}{56.04}}
& \multicolumn{3}{c}{82.51\ci{76.02}{89.00}}
& \multicolumn{3}{c}{52.99\ci{51.04}{54.94}}
& \multicolumn{3}{c}{1.45} \\

\midrule
\multicolumn{13}{c}{\textbf{\textit{Conventional OCR Baselines (non-generative, prompt-independent)}}} \\
\cmidrule(lr){1-13}
& \multicolumn{3}{c}{\textbf{CER}\textcolor{green}{$\downarrow$}}
& \multicolumn{3}{c}{\textbf{WER}\textcolor{green}{$\downarrow$}}
& \multicolumn{3}{c}{\textbf{1$-$NED}\textcolor{green}{$\uparrow$}}
& \multicolumn{3}{c}{\textbf{Judge}\textcolor{green}{$\uparrow$}} \\
\cmidrule(lr){2-4} \cmidrule(lr){5-7} \cmidrule(lr){8-10} \cmidrule(l){11-13}

EasyOCR
& \multicolumn{3}{c}{48.43}
& \multicolumn{3}{c}{89.26}
& \multicolumn{3}{c}{48.45}
& \multicolumn{3}{c}{1.98} \\

\rev{EasyOCR (FT)}
& \multicolumn{3}{c}{\rev{37.92}}
& \multicolumn{3}{c}{\rev{76.73}}
& \multicolumn{3}{c}{\rev{--}}
& \multicolumn{3}{c}{\rev{--}} \\

Tesseract
& \multicolumn{3}{c}{98.46}
& \multicolumn{3}{c}{110.05}
& \multicolumn{3}{c}{3.88}
& \multicolumn{3}{c}{1.01} \\

Surya
& \multicolumn{3}{c}{191.32}
& \multicolumn{3}{c}{160.48}
& \multicolumn{3}{c}{11.52}
& \multicolumn{3}{c}{1.02} \\

\bottomrule
\end{tabular}}
\caption{\textsc{BanglaWild} leaderboard. Zero-shot VLMs are evaluated under three prompts on $n=1{,}014$ images. \textbf{Bold} marks the best of the two \emph{deployment-realistic} prompts (P1, P2) per (model, metric); ties are bolded jointly. $^{\dagger}$P3 is an \emph{oracle} condition that prepends gold \textsc{Context}/\textsc{Content} metadata unavailable at inference time and reported for reference only. Subscripts in brackets denote 95\% bootstrapped confidence intervals over the test set (1{,}000 resamples). CER and WER may exceed 100\% when models emit long off-target outputs. Judge scores are on a 1--5 scale from an LLM-as-a-Judge protocol (Appendix~\ref{app:judge_prompt}); the primary judge is Gemini 2.5 Pro for all models, while Gemini entries marked with $^{*}$ are re-scored by Claude Sonnet 4.6 as an out-of-family judge to control for in-family bias. CER is computed over Unicode code points after NFC normalization. \rev{EasyOCR (FT) is EasyOCR with its recognizer fine-tuned on the same 1{,}268-image training split as the LoRA models (Appendix~\ref{app:training}); its 1$-$NED and judge scores were not computed.}}
\label{tab:leaderboard}
\vspace{-3mm}
\end{table*}

\paragraph{A persistent gap despite larger models.}
Tab.~\ref{tab:leaderboard} shows that the best system on \textsc{BanglaWild}, Gemini 2.5 Flash, achieves \textbf{CER 14.08} and WER 29.97 under P1, its best deployment-realistic prompt (P3 reaches CER 13.93 but is an oracle condition and excluded from headline comparisons; see \S\ref{sec:zero_shot}). Performance is not monotonic with model size: Gemini 2.5 Flash outperforms the larger Gemini 2.5 Pro (CER 14.08 vs.\ 21.65; WER 29.97 vs.\ 36.42), while within Qwen3.5, the 9B model \del{slightly surpasses the 27B dense and 35B-A3B MoE variants} \rev{outperforms the 27B dense variant on every metric and roughly matches the 35B-A3B MoE variant (CER 20.38 vs.\ 20.23)}. As the 9B dense, 27B dense, and 35B-A3B MoE models differ architecturally rather than forming a clean scaling series, this non-monotonicity should be interpreted cautiously. InternVL3.5 and GPT-5.4-nano collapse across scales (CER $>$100), often producing outputs longer than the ground truth. These flat or inverse scaling trends suggest that Bengali scene text recognition on \textsc{BanglaWild} depends more on training data composition than model size.
\paragraph{Prompt language has a small but systematic effect on the mid-tier.}
Bengali-instructed P2 improves over English P1 on CER for 7 of 15 models, with the largest gains in the mid-tier (Claude Sonnet 4.6: $\Delta\text{CER}=-4.69$; Claude Haiku 4.5: $-8.30$; Gemini 2.5 Pro: $-3.20$; GPT-5.4-nano: $-54.15$). The effect reverses for the weakest models: InternVL3.5-4B degrades from 118.44 to 156.92 and InternVL3.5-2B from 132.73 to 176.70 under P2, with WER showing the same trend, suggesting these models transcribe the Bengali instruction instead of treating it as a directive, thereby lengthening outputs and inflating both metrics. P3, the oracle metadata condition, achieves the best CER for 9 of 15 models \del{(and every aggregate metric for the weakest open models)} but \del{rarely} \rev{never} the best $1\!-\!$NED, indicating that gold scene priors mainly help models commit to a transcription target rather than improve semantic fidelity. Because this metadata is unavailable at inference, P3 is reported for reference only.
\paragraph{Fine-tuning rescues catastrophic failures but does not raise the ceiling.}
LoRA on the 50\% training split yields large CER reductions for the weakest open-source models: Gemma-3-4B-it drops from $134.82\rightarrow50.72$ ($-62\%$ relative), InternVL3.5-8B from $116.37\rightarrow45.81$ ($-61\%$), and the two smallest InternVL variants from $118$--$133$ to the low 50s ($-56\%$, $-61\%$). By contrast, the strongest open model, Qwen3.5-9B, slightly regresses ($20.38\rightarrow21.53$), while Gemma-3-12B-it improves on CER ($63.89\rightarrow28.91$) and $1{-}$NED ($50.47\rightarrow72.79$) but still trails zero-shot Qwen3.5-9B. The key finding: LoRA on 1{,}268 images mitigates catastrophic failures (refusals, off-target English captions, and cross-script contamination) rather than improving the Bengali recognition already achieved by competent models. \rev{Fine-tuning EasyOCR's recognizer on the same split lowers its CER from $48.43\rightarrow37.92$ ($-21.7\%$ relative) and WER from $89.26\rightarrow76.73$ ($-14.0\%$), but it remains far behind zero-shot Gemini~2.5 Flash (CER 14.08), and its residual errors stay visual-dominant (\S\ref{sec:error}).}
\paragraph{Edit-distance and semantic judgments diverge at the top.}
At the top of the leaderboard, CER and the LLM judge disagree: Claude Sonnet 4.6 achieves the highest judge score across all fifteen systems (3.82 at P2) despite trailing Gemini 2.5 Flash on CER by about four points (18.46 vs.\ 14.08 at each model's best deployment-realistic prompt). Gemini 2.5 Pro (judge 3.67) ranks between Flash (3.69) and Qwen3.5-9B (3.30), again differing from the CER ordering. As discussed in \S\ref{sec:error}, Sonnet appears to produce more spelling-normalized yet meaning-preserving transcriptions, which edit distance penalizes but the judge does not. So, we treat the judge score as complementary to CER rather than a standalone ranking metric.
\paragraph{Cross-family judge validation isolates in-family bias.}
The primary judge in Tab.~\ref{tab:leaderboard} is Gemini~2.5 Pro, which evaluates all fifteen systems under the rubric in Appendix~\ref{app:judge_prompt}. Because Gemini~2.5 Flash and Gemini~2.5 Pro are judged by a model from the same family, prior work on self-preference, family bias, and preference leakage \citep{panickssery2024llm,wataoka2024selfpref,spiliopoulou2025playfavorites,li2025preference} suggests their scores may be inflated. To quantify this, we re-scored both Gemini models with Claude Sonnet~4.6 using the identical rubric. Gemini~2.5 Flash drops from $\{3.84, 3.86, 3.86\}$ (Gemini Pro judge; P1/P2/P3) to $\{3.68, 3.69, 3.70\}$ (Sonnet judge), and Gemini~2.5 Pro from $\{3.68, 3.72, 3.75\}$ to $\{3.58, 3.67, 3.63\}$, implying in-family inflation of $0.16$--$0.17$ points for Flash and $0.05$--$0.12$ for Pro on the 1--5 scale. We therefore report the Sonnet-judged scores for both Gemini models in Tab.~\ref{tab:leaderboard}; the remaining thirteen models retain Gemini Pro scores because full cross-family re-scoring was beyond our budget. This makes the CER-judge inversion at the top conservative: correcting Sonnet~4.6 for its own potential family bias would require an out-of-family Sonnet judge, so its reported semantic advantage is more likely understated than overstated. More generally, judge-based comparisons between Gemini and non-Gemini models should be interpreted with possible residual in-family bias, and we treat judge scores as complementary to CER rather than a standalone ranking metric.
\paragraph{Where the errors live.}
Tabs.~\ref{tab:bangla_wild_results}, \ref{tab:bangla_wild_attributes}, and~\ref{tab:per_category_all_prompts} characterize the failure surface. \emph{Marginals (Tab.~\ref{tab:bangla_wild_results}):} on the Context axis (Gemini~2.5 Flash, P1), CER increases from \textsc{Printed Material}~(7.70) to \textsc{Packaging}~(10.26), \textsc{Signboard}~(13.69), \textsc{Wall}~(13.74), and \textsc{Vehicle}~(30.35), a $3.9\times$ span separating planar from hand-painted curved surfaces. On the Content axis, CER rises from \textsc{Instructional}~(8.76) to \textsc{Religious}~(11.94), \textsc{Advertisement}~(12.42), \textsc{Creative}~(13.39), \textsc{Political}~(17.42), and \textsc{Colloquial}~(21.69), a $2.5\times$ span reflecting increasing typographic variability. Both rankings are \rev{broadly} consistent across Claude Sonnet~4.6 and Qwen3.5-9B, indicating dataset rather than model effects. \emph{Joint cells (Tab.~\ref{tab:per_category_all_prompts}):} the marginals compound. The easiest cells are \textsc{Printed Material}$\times$\textsc{Instructional} (CER~2.33) and \textsc{Signboard}$\times$\textsc{Instructional} (5.81); the hardest are \textsc{Vehicle}$\times$\textsc{Creative} (48.22--60.38), \textsc{Vehicle}$\times$\textsc{Colloquial} (33.03--47.48), and \textsc{Wall}$\times$\textsc{Colloquial} (30.05--34.91). The effects are superadditive: \textsc{Vehicle}$\times$\textsc{Creative} (60.38 at P1) exceeds the sum-minus-mean prediction from the marginals, showing that difficulty arises from the interaction of hand-painted surfaces and informal language rather than either factor alone. \emph{Diagnostic axes (Tab.~\ref{tab:bangla_wild_attributes}):} \textsc{Curved Text} increases CER by $2.2\times$ for both Gemini~2.5 Flash ($13.07\rightarrow28.69$) and Qwen3.5-9B ($18.88\rightarrow41.98$); \textsc{Heavy Occlusion} by $1.6\times$ and $1.3\times$; and \textsc{Cluttered Background} by $1.8\times$ and $1.4\times$, respectively. In contrast, \textsc{Artistic} versus \textsc{Printed} fonts increase CER by only $1.4\times$ and $1.2\times$. The dominant challenges are layout non-linearity and communicative informality, rather than scene text augmentation targets such as perspective, occlusion, or decorative typography. \rev{Because the annotation protocol is identical across all cells, these within-benchmark contrasts (up to $26\times$ between joint cells) locate the difficulty in the images themselves rather than in our protocol.}

\vspace{-1mm}
\section{Error Analysis}
\label{sec:error}
\vspace{-1mm}

\begin{table*}[t]
\centering
\scriptsize
\setlength{\tabcolsep}{3pt}
\renewcommand{\arraystretch}{1.05}
\resizebox{1.0\textwidth}{!}{
\begin{tabular}{l cccc *{12}{c}@{}}
\toprule
\multirow{2}{*}{\textbf{Error type}}
& \multicolumn{4}{c}{\textbf{Conventional OCR}}
& \multicolumn{3}{c}{\textbf{Gemini-2.5-Flash}} & \multicolumn{3}{c}{\textbf{Qwen3.5-9B}}
& \multicolumn{3}{c}{\textbf{Claude Sonnet 4.6}} & \multicolumn{3}{c}{\textbf{Gemma-3-12B-it}} \\
\cmidrule(lr){2-5} \cmidrule(lr){6-8} \cmidrule(lr){9-11} \cmidrule(lr){12-14} \cmidrule(lr){15-17}
& Tesseract & EasyOCR & \rev{EasyOCR (FT)} & Surya & P1 & P2 & P3 & P1 & P2 & P3 & P1 & P2 & P3 & P1 & P2 & P3 \\
\midrule
\multicolumn{17}{l}{\rev{\textit{(a) Word-level events (one per misaligned word)}}} \\
Lexical word substitution
& 11.10 & 26.56 & \rev{28.19} & 16.09 & 26.9 & 27.5 & 27.2 & 24.8 & 24.1 & 24.3 & 21.95 & 23.59 & 23.35 & 43.52 & 41.11 & 43.14 \\
Visual grapheme misread
& 2.37 & 10.11 & \rev{17.13} & 4.88 & 18.9 & 17.3 & 18.7 & 22.3 & 22.8 & 23.0 & 17.50 & 17.35 & 17.43 & 13.71 & 11.94 & 13.79 \\
Word-boundary segmentation
& 17.43 & 27.78 & \rev{26.99} & 11.33 & 17.3 & 19.3 & 16.5 & 11.7 & 13.0 & 12.2 & 14.45 & 15.89 & 13.93 & 13.52 & 18.32 & 13.39 \\
Multi-character word corruption
& 1.67 & 10.34 & \rev{6.73} & 2.93 & 10.4 & 10.2 & 10.3 & 12.6 & 12.5 & 12.5 & 9.41 & 9.21 & 9.23 & 11.08 & 10.77 & 11.40 \\
Vowel-sign (matra) substitution
& 0.23 & 4.25 & \rev{6.87} & 0.30 & 7.5 & 6.8 & 7.2 & 7.2 & 7.4 & 7.1 & 6.41 & 6.02 & 6.01 & 4.19 & 3.76 & 4.47 \\
Unicode normalization
& 0.40 & 2.01 & \rev{1.49} & 1.83 & 4.8 & 5.3 & 5.6 & 11.0 & 10.7 & 11.0 & 16.58 & 16.26 & 16.63 & 1.53 & 0.95 & 1.51 \\
Consonant homophone
& 0.03 & 1.16 & \rev{0.39} & 0.12 & 2.2 & 2.4 & 2.2 & 1.0 & 0.9 & 1.0 & 1.20 & 1.38 & 1.48 & 0.34 & 0.36 & 0.29 \\
Conjunct / hasanta error
& 0.06 & 4.63 & \rev{2.13} & 0.06 & 1.4 & 1.4 & 1.6 & 1.0 & 1.1 & 1.0 & 1.25 & 1.14 & 1.28 & 0.26 & 0.24 & 0.30 \\
Compound orthographic
& 0.06 & 1.34 & \rev{1.09} & 0.00 & 1.0 & 1.1 & 0.9 & 0.9 & 0.8 & 0.9 & 0.76 & 0.77 & 0.82 & 0.69 & 0.63 & 0.78 \\
Nasal diacritic error
& 0.00 & 1.20 & \rev{0.38} & 0.00 & 0.7 & 0.6 & 0.7 & 0.3 & 0.4 & 0.3 & 0.71 & 0.72 & 0.76 & 0.11 & 0.10 & 0.18 \\
\rev{\quad Subtotal (\% of all events)}
& \rev{33.35} & \rev{89.38} & \rev{91.39} & \rev{37.54} & \rev{91.1} & \rev{91.9} & \rev{90.9} & \rev{92.8} & \rev{93.7} & \rev{93.3} & \rev{90.22} & \rev{92.33} & \rev{90.92} & \rev{88.95} & \rev{88.18} & \rev{89.25} \\
\midrule
\multicolumn{17}{l}{\rev{\textit{(b) Image-level events (one per image)}}} \\
Script contamination
& 1.50 & 1.01 & \rev{0.79} & 51.83 & 6.8 & 5.5 & 6.7 & 5.6 & 4.8 & 5.4 & 8.23 & 6.18 & 7.27 & 8.57 & 10.17 & 8.38 \\
Semantic hallucination
& 2.32 & 4.38 & \rev{3.47} & 2.62 & 0.6 & 0.9 & 0.8 & 1.0 & 1.0 & 0.7 & 1.23 & 1.01 & 1.31 & 2.06 & 1.44 & 1.96 \\
Numeric / temporal error
& 0.03 & 1.03 & \rev{0.79} & 0.00 & 1.2 & 1.2 & 1.3 & 0.3 & 0.4 & 0.3 & 0.27 & 0.34 & 0.33 & 0.14 & 0.15 & 0.11 \\
Truncation / no output
& 62.80 & 3.16 & \rev{2.79} & 7.95 & 0.1 & 0.1 & 0.1 & 0.1 & 0.1 & 0.2 & 0.03 & 0.05 & 0.05 & 0.26 & 0.06 & 0.29 \\
Spurious extra content
& 0.03 & 1.04 & \rev{0.77} & 0.06 & 0.1 & 0.2 & 0.1 & 0.1 & 0.1 & 0.1 & 0.03 & 0.08 & 0.11 & 0.02 & 0.02 & 0.02 \\
\rev{\quad Subtotal (\% of all events)}
& \rev{66.68} & \rev{10.62} & \rev{8.61} & \rev{62.46} & \rev{8.8} & \rev{7.9} & \rev{9.0} & \rev{7.1} & \rev{6.4} & \rev{6.7} & \rev{9.79} & \rev{7.66} & \rev{9.07} & \rev{11.05} & \rev{11.84} & \rev{10.76} \\
\midrule
\rev{Visual-consistent (\% of word-level)}
& \rev{45.40} & \rev{52.60} & \rev{56.95} & \rev{63.67} & \rev{61.7} & \rev{59.8} & \rev{61.8} & \rev{64.3} & \rev{63.4} & \rev{64.1} & \rev{54.16} & \rev{54.32} & \rev{55.00} & \rev{76.80} & \rev{72.37} & \rev{76.56} \\
\rev{Orthographic (\% of word-level)}
& \rev{1.14} & \rev{14.07} & \rev{11.88} & \rev{1.28} & \rev{14.1} & \rev{13.4} & \rev{13.9} & \rev{11.2} & \rev{11.3} & \rev{11.0} & \rev{11.45} & \rev{10.86} & \rev{11.38} & \rev{6.28} & \rev{5.77} & \rev{6.75} \\
Exact-match accuracy (\%)
& 0.16 & 4.69 & \rev{5.03} & 0.63 & 35.6 & 35.9 & 35.3 & 16.1 & 16.3 & 16.1 & 25.81 & 26.09 & 26.60 & 4.42 & 4.54 & 5.41 \\
\bottomrule
\end{tabular}}
\caption{Unified error taxonomy on \textsc{BanglaWild} for three conventional OCR baselines\rev{, EasyOCR fine-tuned on our training split (FT),} and four VLM families across the three prompts. OCR values are \% of residual errors and are prompt-independent. VLM values are \% of errors per prompt. \rev{Panel (a) lists word-level events (one per misaligned word) and panel (b) image-level events (one per image); each subtotal is that level's share of all events. The last block re-expresses the visual-consistent (lexical, grapheme, multi-character) and orthographic (matra, homophone, conjunct, compound, nasal) categories as shares of word-level events.} A detailed explanation of the error classification protocol can be found in Appendix~\ref{app:classifier}.}
\label{tab:error_taxonomy}
\vspace{-4mm}
\end{table*}

Aggregate CER/WER quantify \emph{how far} predictions deviate from ground truth, but not \emph{how} they fail. We therefore classify every error into fifteen mutually exclusive categories (Tab.~\ref{tab:error_taxonomy}) using a deterministic classifier applied to four VLM families (closed-source Gemini-2.5-Flash and Claude Sonnet~4.6, open-source Qwen3.5-9B and Gemma-3-12B-it) and three OCR baselines (Tesseract, EasyOCR, Surya)\rev{, plus EasyOCR after fine-tuning}. \rev{The classifier sees only the ground-truth and predicted strings, so its categories describe the form of an error, not where in a model it arises. We report word-level events (one per misaligned word) and image-level events (one per image) as separate panels of Tab.~\ref{tab:error_taxonomy}.} Unless otherwise noted, the analysis focuses on Gemini-2.5-Flash and Qwen3.5-9B as representative strong systems; the full per-model breakdown is given in Tab.~\ref{tab:error_taxonomy}.

\paragraph{Bengali scene text failure is dominated by \rev{errors consistent with} visual mis-recognition, not orthographic confusion.}
Lexical substitution, single-character visual misread, and multi-character word corruption\del{, all reflecting \emph{vision-encoder} failures,}\rev{, whose string patterns are consistent with visual mis-recognition,} account for \del{$\sim\!56\%$ of Gemini-2.5-Flash errors and $\sim\!60\%$ of Qwen3.5-9B errors} \rev{60--62\% of Gemini-2.5-Flash and 63--64\% of Qwen3.5-9B word-level errors} across all prompts. In contrast, the \emph{orthographic} categories emphasized in prior Indic OCR work (matra substitution, consonant homophones, conjunct/hasanta errors, nasal diacritics, and compound orthographic edits) contribute only \del{$\sim\!13\%$ and $\sim\!10\%$} \rev{13--14\% and $\sim\!11\%$}, yielding visual: orthographic ratios of \del{$4.4\!\times\!$ and $5.8\!\times\!$} \rev{$4.4$--$4.5\times$ and $5.6$--$5.8\times$}, respectively. This reverses the prevailing assumption in Bengali OCR: for generative VLMs, \del{the bottleneck is no longer decoding complex orthography but \emph{recognizing glyphs} on cluttered, hand-painted, or oblique surfaces} \rev{the remaining errors look far less like failures to decode complex orthography than like failures to \emph{recognize glyphs} on cluttered, hand-painted, or oblique surfaces}. The qualitative errors are consistent with visual rather than linguistic failures: \banglafont{ফ্ল্যাটটি} (\textit{phlyāṭṭi}) $\to$ \banglafont{ফ্যাক্টরি} (\textit{phyākṭari}) and \banglafont{কপাট} (\textit{kapāṭa}) $\to$ \banglafont{লোহ} (\textit{loha}) are neither homophones, near-conjuncts, nor plausible language-prior substitutions, \del{but} \rev{and instead resemble} visual misreadings. \del{This pattern is architecture-independent: on} \rev{The same profile appears in} EasyOCR, the only conventional baseline that reliably reads Bengali\del{, visual errors}\rev{: errors consistent with visual mis-recognition} outnumber orthographic ones by $3.7\times$ \rev{zero-shot and by $4.8\times$ after fine-tuning on our training split} (Tab.~\ref{tab:error_taxonomy}), \del{matching the VLMs} \rev{comparable to Gemini-2.5-Flash and Qwen3.5-9B}. \del{The bottleneck therefore lies in the in-the-wild imagery rather than any specific model class.} \rev{Tesseract and Surya fail before their profiles become informative (62.8\% of Tesseract's events are truncations and 51.8\% of Surya's are script contamination), so we do not extend this observation to conventional OCR in general.}

\paragraph{The classical conjunct bottleneck has largely closed.}
Conjunct/hasanta errors account for only $1.0$--$1.6\%$ of errors, despite CRNN-era Bengali OCR treating \textit{yuktakshara} recognition as the primary structural challenge and prior Indic benchmarks \citep{nayef2019icdar, gunna2022transferlearning} reporting only aggregate metrics that do not isolate conjunct errors. For modern VLMs, conjunct structure thus accounts for only a small share of the observed errors. Even on EasyOCR, conjunct/hasanta errors are more frequent ($4.63\%$ of residual errors, a \~{}$3\text{-}4\times$ higher share than VLMs) but remain secondary to segmentation and lexical substitution\rev{, and fine-tuning roughly halves their share ($2.13\%$)}. \del{The bottleneck has therefore diminished across model classes, not just VLMs.} \rev{Conjunct errors are thus a minor share of residual errors for both the VLMs and EasyOCR.}

\paragraph{Prompting mainly affects cross-script contamination.}
Among the fifteen error categories, Script contamination is consistently affected by prompt language: it drops from $6.8\%\to5.5\%$ on Gemini and $5.6\%\to4.8\%$ on Qwen under P2, then rebounds at P3 when English is reintroduced. Lexical substitution, grapheme misread, matra error, and conjunct error remain stable within $\pm1.5$pp across P1/P2/P3 for both models. This suggests prompt language \del{gates the \emph{output language head}} \rev{mainly affects the output language} (Bengali conditioning reduces probability mass on residual Devanagari or Latin transliteration pathways) but \del{not the visual encoder, where the dominant errors occur. This supports the visual-bottleneck claim above and bounds} \rev{leaves the dominant, visually consistent errors unchanged, which limits} what prompt engineering alone can achieve on this benchmark.
\vspace{-2mm}
\paragraph{Tokenization and Unicode artefacts inflate WER without reducing readability, and vary by model.}
Word-boundary segmentation (12--19\% of errors) and Unicode/punctuation normalization (5--11\%) together account for 22--25\% of all errors while leaving readable content unchanged: an attached \banglafont{এনা} (\textit{en\=a}) $\rightarrow$ \banglafont{এন} (\textit{ena}) \banglafont{এ} (\textit{e}), a typographic \textquotedblright{} vs.\ \textquotedblleft{}, or an NFC-equivalent code point. These categories explain much of the Gemini-Qwen gap in aggregate metrics: Qwen3.5-9B produces 2.3$\times$ more Unicode-normalization noise than Gemini-2.5-Flash ($\sim$11\% vs.\ $\sim$5\%), despite visually identical output. They also explain the Sonnet-Flash judge inversion in~\S\ref{sec:results}: Sonnet's higher judge score is consistent with more spelling-normalized yet meaning-preserving transcriptions, which edit-distance metrics penalize but the judge does not. WER-based deployment decisions for Bengali scene text should therefore account for each model's share of cosmetic errors.

\paragraph{Two small categories carry disproportionate deployment risk.}
Semantic hallucination ($0.6\text{-}1.0\%$) and numeric/temporal substitution ($0.3\text{-}1.3\%$) are individually rare, but uniquely produce fluent, plausible Bengali that cannot be recognized as incorrect from the output alone. Hallucinations preserve discourse type while replacing lexical identity: \banglafont{ছাত্র ইউনিয়ন} (\textit{ch\=atra iuniyana}) $\rightarrow$ \banglafont{শুভ জন্মদিন} (\textit{\'subha janmadina}), \banglafont{হর্ণ} (\textit{har\d{n}a}) $\rightarrow$ \banglafont{শুভ নববর্ষ} (\textit{\'subha nababar\d{s}a}). Numeric substitutions silently rewrite prices, years, and phone numbers (\banglafont{১৪৩৩} (1433) $\rightarrow$ \banglafont{১৪৩০} (1430), \banglafont{২০২৫} (2025) $\rightarrow$ \banglafont{২০২৩} (2023)) and are $\sim\!4\times$ more common in Gemini-2.5-Flash than Qwen3.5-9B (1.2\% vs.\ 0.3\%), with rates unchanged across prompts, suggesting a training-time prior over plausible Bengali calendar years rather than a prompt effect. Although neither category dominates CER, they matter most for verbatim applications such as transcribing addresses, receipts, election material, and medical packaging. They also explain cases where CER is low but LLM-judge scores are poor: the judge detects semantic mismatches against the reference, whereas users see only fluent but incorrect text. Additional labelled examples are provided in Appendix~\ref{app:error}.
\vspace{-2mm}
\paragraph{Per-word comparisons reveal shared orthographic confusions and model-specific hallucinations.}
Tab.~\ref{tab:mismatch} compares ground-truth Bengali words, their corrected forms, and the outputs of four leading systems on identical inputs. Two patterns emerge: 1. Most disagreements reflect orthographic contrasts subtle to non-Bengali readers: retroflex versus dental nasal (\banglafont{ণ}/\banglafont{ন}; \textit{ṇ}/\textit{n}), long versus short vowel signs (\banglafont{ী}/\banglafont{ি}; \textit{ī}/\textit{i}), retroflex versus dental sibilant (\banglafont{ষ}/\banglafont{স}; \textit{ṣ}/\textit{s}), and aspirated versus unaspirated retroflex (\banglafont{ঢ়}/\banglafont{ড়}; \textit{ḍh}/\textit{ṛ}). Although CER penalizes these equally, models diverge: for \banglafont{সুবিদা} (\textit{subidā}), Qwen3.5-35B-A3B alone matches the ground truth, Qwen3.5-9B and Claude Sonnet 4.6 produce the corrected \banglafont{সুবিভা} (\textit{subibhā}), and Gemini 2.5 Flash outputs a third variant, \banglafont{সুবিধা} (\textit{subidhā}). 2. Hallucinations are model-specific: for~\banglafont{হর্ণ} (\textit{harṇ}), Qwen3.5-9B outputs \textit{jana}, Qwen3.5-35B-A3B outputs \textit{tarka}, Gemini 2.5 Flash hallucinates \banglafont{শুভ নববর্ষ} (\textit{śubha nababarṣa}), and Claude Sonnet 4.6 produces the corrected \banglafont{হর্ন} (\textit{harn}), a textbook correction error revealed by the dual annotation in \S\ref{sec:dataset}. The diversity of outputs on identical inputs suggests distinct language-prior pathways rather than shared visual confusion.
\definecolor{bestblue}{HTML}{BFDBFE}

\section{Conclusion}
\label{sec:conclusion}
We introduced \textsc{BanglaWild}, the first in-the-wild Bengali scene text recognition benchmark to pair dual verbatim/standard annotation with evaluation of both conventional OCR and generative VLMs on the same data. Across 15 VLMs, 3 conventional OCR systems, 3 prompting strategies, 6 LoRA-adapted variants, \rev{a fine-tuned EasyOCR,} and a 15-class error taxonomy, our key finding inverts a long-standing CRNN-era assumption: \rev{word-level errors consistent with} visual mis-recognition, not orthographic confusion, \del{dominates Bengali scene text recognition failure} \rev{dominate the observed distribution}, a pattern that holds \del{across VLMs and a conventional OCR baseline} \rev{in four VLM families and in EasyOCR before and after fine-tuning}, and the classical conjunct bottleneck has largely closed. The dual annotation additionally surfaces generative-model-specific failures, silent over-correction and fluent hallucination, that conventional OCR structurally cannot exhibit. We hope our dataset, taxonomy, and findings will foster future research toward more visually robust Indic-script recognition systems.
\section*{Limitations}
\label{sec:limitations}
Despite the contributions of this work, several limitations remain. \rev{\textsc{BanglaWild} (2{,}535 images of five surface types, all from Bangladesh) omits settings such as digital displays and receipts, as well as conventions of other Bengali-speaking regions.} \del{The benchmark covers Bengali text only; code-mixed
Bengali--English images were excluded at the filtering stage to
keep the evaluation target unambiguous, leaving the
code-mixing problem to future purpose-built resources.} \rev{Filtering removed 675 low-quality, unreadable, code-mixed, or offensive images; such inputs are common in deployment and likely harder, so our error rates are probably optimistic, and we did not evaluate models on them, leaving code-mixed Bengali--English text to future purpose-built resources. Capture conditions (resolution, blur, exposure, low light) were not recorded, so some apparent visual mis-recognitions may reflect image quality rather than the script; isolating these factors is future work.} The categorical distribution reflects natural abundance rather
than balance: \textsc{Wall} dominates \textsc{Context} and
six of the thirty possible (\textsc{Context},
\textsc{Content}) combinations contain no images, so per-cell
results in the sparser categories carry higher variance than
the marginals suggest\rev{, as do over-correction estimates from the 87 verbatim/standard images}. \del{Our LoRA experiments adapt six
open-source models on $\sim$1{,}268 training images and use
a single random seed under our compute budget; the negative
result on already-competent models tests targeted small-scale
adaptation, not whether full-scale pre-training revision would
help.} \rev{Fine-tuning uses $\sim$1{,}268 training images and one run per model, so the negative LoRA result on competent models concerns small-scale adaptation, not full-scale pre-training. The conventional-OCR comparison rests on EasyOCR, the only such system that reads Bengali reliably; a modern dedicated scene-text pipeline would be a stronger test. We do not compare against another real-world Bengali or Indic benchmark, where differing annotations and metrics would confound results; our evidence on difficulty is within-benchmark (\S\ref{sec:results}).} \del{The LLM-as-a-Judge is itself a frontier VLM and may share blind spots 
with the systems it scores, particularly leniency toward 
fluent-but-incorrect Bengali; we validate against native-annotator 
re-scoring ($\rho = 0.89$). To mitigate in-family bias 
\mbox{\citep{panickssery2024llm,li2025preference}}, we re-scored both 
Gemini-family systems with Claude Sonnet 4.6 as an out-of-family 
judge and report those values in \mbox{Tab.~\ref{tab:leaderboard}}; 
budget constraints prevented out-of-family re-scoring of the 
remaining thirteen models, so residual in-family inflation cannot 
be excluded for any non-Gemini cell. We report judge scores 
alongside, not in place of, edit-distance metrics.} \rev{The LLM judge may share blind spots with the systems it scores, such as leniency toward fluent but incorrect Bengali; despite strong agreement with native annotators ($\rho = 0.89$) and out-of-family re-scoring of both Gemini systems, residual in-family bias \citep{panickssery2024llm,li2025preference} cannot be excluded for the other thirteen models, so judge scores complement rather than replace edit-distance metrics. Our error classifier sees only the ground-truth and predicted strings, so its categories describe the form of an error, not its cause: errors consistent with visual mis-recognition could also arise from multimodal alignment or the decoder's language prior, and we did not check category assignments against the images.} Finally, \del{our fifteen-class error taxonomy was
applied to four VLM families (Gemini 2.5 Flash, Qwen3.5-9B,
Claude Sonnet 4.6, and Gemma-3-12B-it) and three conventional
OCR baselines (Tesseract, EasyOCR, Surya) using an identical
deterministic classifier \mbox{(Appendix~\ref{app:classifier})}; the
detailed per-category analysis in \mbox{\S\ref{sec:error}} focuses on
Gemini 2.5 Flash and Qwen3.5-9B as representative strong
systems, and} the reported \rev{taxonomy} percentages should not be assumed to
transfer without re-classification to systems with substantively
different failure profiles.

\section*{Ethical Statement}
\label{sec:ethics}
This work involves the creation and use of a Bengali
scene text benchmark sourced from public-facing physical
surfaces. Trained collaborators captured all images in
public space consistent with local norms for street
photography; the benchmark targets text, not people, and
where human faces or vehicle registration plates appeared
incidentally in the frame, we blurred them and verified that
the remaining content is non-uniquely-identifying prior to
release. Vulgar, offensive, and entirely unreadable content
was excluded at the filtering stage. We retain politically
sensitive material such as protest slogans, election posters, and
religious verses, because excluding it would systematically
misrepresent the visual environment the benchmark is intended
to characterize; such content is transcribed factually and
without endorsement. Annotators from Bangladesh (three native Bengali-speaking transcribers, one expert verifier, and two additional annotators for judge validation) were compensated at standard per-sample rates and consented to the release of
their adjudicated annotations. We acknowledge that VLMs
evaluated on this benchmark may produce inaccurate, biased, or
culturally inappropriate outputs; the benchmark exists to
surface such failures, not to validate deployment. The failure modes
we document carry concrete downstream risks if such systems are deployed
for verbatim Bengali transcription: misread addresses can cause
misdelivery, silently altered numerals can corrupt prices, dates, and
dosages on medical packaging, and hallucinated or over-corrected text on
political and religious surfaces can misattribute or distort sensitive
content. These risks fall disproportionately on a low-resource-language
population with limited recourse, which is precisely why we release
\textsc{BanglaWild} \del{as a held-out diagnostic resource rather than a
training corpus} \rev{primarily as a diagnostic resource whose training split supports small-scale adaptation studies, not commercial training}; the \rev{data} release license (CC BY-NC 4.0) reflects this intended
use. \rev{Our code is released under the MIT License.}

\section*{Acknowledgments}
\rev{We thank the annotators who transcribed and verified the \textsc{BanglaWild} images, and the anonymous reviewers and area chair for their constructive feedback.}



\bibliography{custom}

\appendix


\section{Category Definitions}
\label{app:categories}

While the main text gives informal descriptions of each category for the reader, this section formally defines each value of the \textsc{Context} and \textsc{Content} axes to support reproducibility and annotation extension.

\subsection{Context (5 values)}
\begin{itemize}\itemsep0pt
  \item \textbf{Vehicle.} Text painted, decaled, or otherwise affixed to buses, trucks, three-wheelers, and similar conveyances. Typically rendered in stylised hand-painted vehicle lettering with strong artistic flourishes.
  \item \textbf{Signboard.} Shop and storefront signage, including hand-painted, vinyl-printed, and back-lit varieties. Text is the dominant visual element of the surface.
  \item \textbf{Packaging.} Consumer product packaging, predominantly processed food and household goods. Text is typically small, dense, and embedded in busy commercial designs.
  \item \textbf{Wall.} Text appearing on built surfaces such as painted notices, posters affixed to walls, stencilled markings, and free-painted wall-text. The surface itself is not portable.
  \item \textbf{Printed Material.} Banners, leaflets, and posters that are free-standing, strung, or hand-held, rather than affixed to a wall surface.
\end{itemize}

\subsection{Content (6 values)}
\begin{itemize}\itemsep0pt
  \item \textbf{Religious.} Verses, invocations, names, and aphorisms drawn from religious traditions. \textsc{BanglaWild} includes content from Islamic, Hindu, Buddhist, and Christian traditions.
  \item \textbf{Political.} Slogans, election material, and protest-related text. Identification is based on overtly political content rather than on the surface or the speaker.
  \item \textbf{Advertisement.} Commercial promotions for products and services, including food, construction materials, telecommunications, and consumer goods.
  \item \textbf{Instructional.} Directive or informational notices in public space: prohibitions, warnings, requests for assistance, and traffic/safety instructions.
  \item \textbf{Creative.} Song lyrics, poetry excerpts, literary quotations, and aphorisms appearing in public visual culture (rather than in printed books).
  \item \textbf{Colloquial.} Everyday, miscellaneous public-space text not fitting the other categories, such as names, greetings, and idiosyncratic markings. Defined explicitly as a residual category so that ambiguous text is not forced into ill-fitting labels.
\end{itemize}

The six \textsc{Content} categories are mutually exclusive. Where an image straddles two values (e.g., a religious aphorism deployed as a political slogan), annotators were instructed to select the value that better captures the speaker's primary communicative intent and to flag the image for adjudication.

\section{Annotation Rubric}
\label{app:rubric}

We reproduce the rubric given to annotators. The rubric was developed iteratively over a pilot annotation of approximately 200 images before being frozen for the main annotation pass.

\begin{table}[h]
\centering
\small
\begin{tabular}{l c}
\toprule
Diagnostic attribute & Count \\
\midrule
\textsc{Curved Text} - No & 2335 \\
\textsc{Curved Text} - Yes & 200 \\
\midrule
\textsc{Font Style} - Handwritten & 1436 \\
\textsc{Font Style} - Printed & 644 \\
\textsc{Font Style} - Artistic & 455 \\
\midrule
\textsc{Occlusion} - None & 2009 \\
\textsc{Occlusion} - Partial & 503 \\
\textsc{Occlusion} - Heavy & 23 \\
\midrule
\textsc{Background Complexity} - Clean & 2158 \\
\textsc{Background Complexity} - Moderate & 279 \\
\textsc{Background Complexity} - Cluttered & 98 \\
\bottomrule
\end{tabular}
\caption{Distribution of diagnostic attributes in \textsc{BanglaWild}.}
\label{tab:diagnostic_attributes}
\end{table}

\paragraph{Transcription.}
\begin{enumerate}\itemsep0pt
  \item Transcribe verbatim, not normatively. Type exactly what is written in the image, including spelling errors, unconventional spacing, or non-standard vowel sign placement. Do not ``fix'' misspellings, even obvious ones.
  \item If the in-image text is misspelled, additionally fill the \texttt{standard\_form} field with the orthographically standard form. Leave it blank if no deviation is present.
  \item Transcribe only Bengali text. If the image also contains English, numerals, or other-script content, transcribe only the Bengali. Flag the image for the code-mixed exclusion filter if Bengali is not the dominant content.
  \item For multi-line text, transcribe top-to-bottom, left-to-right within each line. For text on curved baselines, follow the baseline in the natural reading direction.
  \item Preserve punctuation as it appears. Use the Bengali danda where present rather than a Latin period.
  \item If uncertain about a character, mark it with \texttt{[?]}; the expert verifier will adjudicate. Do not guess. If the text is entirely unreadable, flag the image for the unreadable-text exclusion filter.
\end{enumerate}

\paragraph{Categorical labels.}
\textsc{Context} answers ``what physical medium does this text appear on?'' and is determined by the surface. \textsc{Content} answers ``what is the text trying to communicate?'' and is determined by intent independent of the surface.

\paragraph{Diagnostic attributes.}
Beyond the categorical axes, each image carries four diagnostic labels targeting known failure modes in scene text recognition:
\textit{Curved Text:} label \textit{Yes} iff the text layout itself is non-linear in the original physical surface; a straight-line sign photographed at an oblique angle remains \textit{No}.
\textit{Font Style:} \textit{Printed} (machine-generated, uniform stroke), \textit{Handwritten} (naturalistic stroke variation), \textit{Artistic} (stylised, decorative, readability traded for aesthetics).
\textit{Occlusion:} \textit{None}, \textit{Partial} (obscured but inferable from context), \textit{Heavy} (recovery difficult even for a native reader).
\textit{Background Complexity:} \textit{Clean} (uniform), \textit{Moderate} (patterned, with text-background separability preserved), \textit{Cluttered} (multiple textures or overlapping objects competing with the text). 
The complete distribution of the diagnostic attributes can be found in Tab. \ref{tab:diagnostic_attributes}

\section{Inter-Annotator Agreement}
Tab.~\ref{tab:iaa} reports the full reliability breakdown referenced in \S~\ref{sec:dataset}.

\begin{table}[t]
\centering
\resizebox{0.65\columnwidth}{!}{
\begin{tabular}{lc}
\toprule
Reliability metric & Score \\
\midrule
Mean pairwise CER (transcription) & 3.2\% \\
Cohen's $\kappa$ \textsc{Context} & 0.93 \\
Cohen's $\kappa$ \textsc{Content} & 0.88 \\
Cohen's $\kappa$ \textsc{Curved Text} & 0.89 \\
Cohen's $\kappa$ \textsc{Font Style} & 0.82 \\
Cohen's $\kappa$ \textsc{Occlusion} & 0.84 \\
Cohen's $\kappa$ \textsc{Background Complexity} & 0.81 \\
\bottomrule
\end{tabular}}
\caption{Inter-annotator agreement on \textsc{BanglaWild}. Conventionally, $\kappa, \alpha \geq 0.80$ indicates strong agreement.}
\label{tab:iaa}
\vspace{-5mm}
\end{table}

\section{Datasheet}
\label{app:datasheet}

We follow the datasheet template of \citet{gebru2021datasheets}; we summarise the essential items here.

\paragraph{Motivation.} \textsc{BanglaWild} was created to enable rigorous evaluation of vision-language models on Bengali scene text under deployment-realistic conditions, since existing Bengali OCR resources are dominated by cropped, rectified, or born-digital text. The benchmark is intended \del{as a held-out evaluation resource, not a training corpus} \rev{primarily as an evaluation resource; its training split supports small-scale adaptation studies such as ours, not the training of commercial systems}. \del{Affiliation and funding details are withheld during double-blind review.} \rev{It was created by researchers at the Islamic University of Technology and BRAC University, Bangladesh, with no external funding.}

\paragraph{Composition.} 2{,}535 image instances, each a JPEG paired with a verbatim gold transcription, \textsc{Context} and \textsc{Content} labels, four diagnostic attribute labels, and, where applicable, an orthographically standard form. The dataset is a directed (non-probabilistic) sample, drawn from a candidate pool of 3{,}210 and filtered as described in \S~\ref{sec:dataset}.

\paragraph{Collection.} Each image was captured directly by a trained collaborator using a consumer mobile-phone camera in the physical setting where the text appeared. Collection spanned different regions of Bangladesh to capture variety in signage conventions, typographic styles and surface types, ensuring that the captured surfaces reflect naturally occurring Bengali scene text rather than the visual environment of any single locality. Camera models and resolutions were not standardised by design. The data captures public surfaces, not people; where individuals appeared incidentally in the frame, faces were blurred and verified non-uniquely-identifying.
 
\paragraph{Preprocessing.} Quality-driven filtering and PII treatment of incidentally visible faces and license plates. No image enhancement, cropping, rectification, or denoising was applied to the released images.

\paragraph{Uses.} Evaluation of Bengali scene text detection, recognition, end-to-end OCR, and VLM visual reasoning over text-bearing images. The diagnostic attributes additionally support targeted robustness studies (e.g., recognition under occlusion, on curved layouts, or against cluttered backgrounds). The dataset should not be used to train commercial OCR or VLM systems.

\paragraph{Sensitive content.} Vulgar and offensive content was excluded at the filtering stage. The \textsc{Political} category contains slogans some readers may find ideologically charged; we retain such content because excluding it would systematically misrepresent the visual environment the benchmark is intended to characterize. No content in the released benchmark depicts violence, hate speech, or targeted harassment.

\paragraph{Distribution \& maintenance.} The dataset \del{will be} \rev{is} released \rev{at} \url{https://github.com/SadabShiper/BanglaWild} under the Creative Commons Attribution-NonCommercial 4.0 International (CC BY-NC 4.0) license for non-commercial research use\rev{; the code, including the error classifier, is released in the same repository under the MIT License}. The current release is \textbf{v1.0}; errata will be issued as point releases preserving backward compatibility of the \texttt{file\_name} and \texttt{Serial} fields. \del{Hosting, maintenance, and contribution channels will be announced in the camera-ready version.} \rev{The authors maintain the repository; errata, questions, and contributions are handled through its issue tracker.}

\section{Prompts}
\label{app:prompts}

\begin{table}[H]
\small
\centering
\begin{tabular}{lp{0.50\columnwidth}}
\toprule
\textbf{Prompt} & \textbf{Description} \\
\midrule
English baseline (\textsc{P1})    & Transcribe-only instruction issued in English. Image-only; deployment-realistic. \\
Bengali baseline (\textsc{P2})     & The \textsc{P1} instruction translated into Bengali script. Image-only; deployment-realistic. \\
Structured metadata (\textsc{P3}) & Gold \textsc{Context} and \textsc{Content} labels prepended to a revised transcription instruction. Oracle condition: the metadata is annotator-provided and unavailable at inference. \\
\bottomrule
\end{tabular}
\caption{Prompting strategies used in the zero-shot benchmark. \textsc{P3} reuses the \textsc{P1} instruction verbatim and only varies the preceding context. \textsc{P1}/\textsc{P2} are image-only and form the deployment-realistic setting; \textsc{P3} is an oracle condition reported for reference and excluded from headline and best-prompt comparisons.}
\label{tab:prompts}
\end{table}

\begin{figure}[H]
\centering
\begin{minipage}{0.95\linewidth}
\begin{tcolorbox}[
    colback=gray!5!white,
    colframe=black,
    boxrule=0.6pt,
    arc=2pt,
    left=6pt, right=6pt, top=6pt, bottom=6pt,
    title=\textbf{P1: Base/Zero-Shot Prompt},
    fonttitle=\small\bfseries
]
\footnotesize

Read and transcribe exactly the text in this image.

\vspace{0.5em}

Output only the text, with no additional commentary, explanation, or formatting.

\end{tcolorbox}
\end{minipage}
\label{fig:p1}
\end{figure}

\begin{figure}[H]
\centering
\begin{minipage}{0.95\linewidth}
\begin{tcolorbox}[
    colback=gray!5!white,
    colframe=black,
    boxrule=0.6pt,
    arc=2pt,
    left=6pt, right=6pt, top=6pt, bottom=6pt,
    title=\textbf{P2: Bengali-Translated Base/Zero-Shot Prompt},
    fonttitle=\small\bfseries
]
\footnotesize

\banglafont{এই ছবিতে যে লেখা আছে তা হুবহু পড়ুন এবং প্রতিলিপি করুন।} (\textit{ei chabite ye lekhā āche tā hubahu paṛuna ebaṃ pratilipi karuna.})
\vspace{0.5em}
\banglafont{শুধুমাত্র লেখাটি লিখুন, কোনো অতিরিক্ত মন্তব্য, ব্যাখ্যা বা ফরম্যাটিং ছাড়া।} (\textit{śudhumātra lekhāṭi likhuna, kono atiriкta mantabya, byākhyā bā pharamyāṭiṃ chāṛā.})

\end{tcolorbox}
\end{minipage}
\label{fig:p2}
\end{figure}

\begin{figure}[H]
\centering
\begin{minipage}{0.95\linewidth}
\begin{tcolorbox}[
    colback=gray!5!white,
    colframe=black,
    boxrule=0.6pt,
    arc=2pt,
    left=6pt, right=6pt, top=6pt, bottom=6pt,
    title=\textbf{P3: Structured Metadata-Augmented Prompt},
    fonttitle=\small\bfseries
]
\footnotesize

This image is annotated with structured metadata:

\vspace{0.4em}

\textbf{Context:} \texttt{\{context\}} \\
\textbf{Content:} \texttt{\{content\}}

\vspace{0.6em}

\textit{Context} refers to the physical location or medium where the text appears.  
\textit{Content} refers to the communicative intent of the text.

\vspace{0.6em}

Based on this information, read and transcribe exactly the text in this image.

\vspace{0.4em}

Output only the text, with no additional commentary, explanation, or formatting.

\end{tcolorbox}
\end{minipage}
\label{fig:p3}
\end{figure}

\begin{figure}[H]
\centering
\begin{minipage}{0.95\linewidth}
\begin{tcolorbox}[
    colback=gray!5!white,
    colframe=black,
    boxrule=0.6pt,
    arc=2pt,
    left=6pt, right=6pt, top=6pt, bottom=6pt,
    title=\textbf{LLM-as-a-Judge Prompt},
    fonttitle=\small\bfseries
]
\footnotesize

You are an expert Bengali (Bengali) evaluator judging transcriptions of stylized in-the-wild Bengali text.

\vspace{0.5em}

\textbf{You will see:}
\begin{itemize}
    \item \textbf{GROUND\_TRUTH} — human-annotated transcription (authoritative)
    \item \textbf{MODEL\_OUTPUT} — system's transcription attempt
\end{itemize}

\vspace{0.3em}

Judge how well \textbf{MODEL\_OUTPUT} matches \textbf{GROUND\_TRUTH}. Accept valid Bengali spelling variants and equivalent Unicode encodings. Ignore whitespace differences.

\vspace{0.5em}

Score on a 1–5 integer scale:
\begin{itemize}
    \item 5 = perfect / fully equivalent
    \item 4 = minor errors, meaning preserved
    \item 3 = several errors, gist recoverable
    \item 2 = major errors, only fragments usable
    \item 1 = wrong, empty, or hallucinated
\end{itemize}

\vspace{0.5em}

Output ONLY a single digit (1, 2, 3, 4, or 5). No other text.

\vspace{0.5em}

\textbf{GROUND\_TRUTH:} \texttt{\{ground\_truth\}} \\
\textbf{MODEL\_OUTPUT:} \texttt{\{model\_output\}}

\end{tcolorbox}
\end{minipage}
\label{fig:p4}
\end{figure}

\section{LLM-as-a-Judge Protocol}
\label{app:judge_prompt}

\paragraph{Motivation.}
CER, WER, and 1$-$NED are surface-level edit-distance
metrics: they penalize every character deviation
uniformly, regardless of whether the deviation preserves
meaning. In Bengali scene text, this is a particularly
acute limitation. A model that emits
\textit{sb\=adh\=ina} (\banglafont{স্বাধীন}) for the ground
truth \textit{sb\=adhina} (\banglafont{স্বাধিন}) differs
in only the long/short \textit{i} vowel sign, yet both
forms denote the same lexeme and a native reader treats
them as equivalent. Conversely, a low-CER output that
silently substitutes a confusable consonant
(\banglafont{ণ}~$\to$~\banglafont{ন}) can shift the lexical
identity entirely. To complement the edit-distance
metrics with a meaning-aware signal, we additionally
score every (ground truth, model output) pair with an
LLM-as-a-Judge protocol.

\paragraph{Judge model and decoding.}
We use Gemini~2.5 Pro as the primary judge for all fifteen systems 
across all three prompts, queried via the Google API with 
\texttt{temperature} = 0 and a \texttt{max\_output\_tokens} cap of 4. 
To control for in-family bias on the two Gemini systems being scored, 
we additionally re-score Gemini~2.5 Flash and Gemini~2.5 Pro with 
Claude Sonnet 4.6 (Anthropic API, identical decoding settings and 
rubric); the Sonnet judged scores replace the Gemini Pro judged ones 
for these two systems in Tab.~\ref{tab:leaderboard}. Cross-family 
re-scoring of the remaining thirteen models was infeasible under our 
compute and API budget (see Limitations). The judge sees only the 
ground-truth transcription and the candidate model output; it does 
not see the source image, the prompting strategy used, or the identity 
of the model being scored. This isolates judgments to textual 
equivalence and prevents any visual-grounding bias from leaking into 
the score.

\paragraph{Rubric.}
The judge assigns an integer score on a 1--5 Likert
scale: 5 denotes fully equivalent transcriptions
(modulo valid Bengali orthographic variants and
Unicode-equivalent encodings), 4 denotes minor errors
that preserve meaning, 3 denotes several errors with
the gist still recoverable, 2 denotes major errors with
only fragments usable, and 1 denotes wrong, empty, or
hallucinated outputs. The judge is instructed to
ignore whitespace differences and to accept canonical
spelling variants, which makes the score robust to the
benign normalization differences that inflate raw CER.

\paragraph{Coverage and aggregation.}
We score the full $n = 1{,}014$ test set for every
(model, prompt) cell reported in
Table~\ref{tab:leaderboard}, yielding
$1{,}014 \times 3$ judgments per zero-shot model and
$1{,}014$ judgments per fine-tuned model. The reported
\textbf{Judge} score is the arithmetic mean of the
per-image scores within each cell. Outputs that
returned a non-digit response from the judge
(\textless~0.1\% of calls across all models) were
re-queried once; any remaining failures were excluded
from the mean for that cell.

\paragraph{Validation.}
To verify that the judge tracks human judgment rather
than amplifying a single model's biases, we sampled 200
(ground truth, model output) pairs stratified across
score bands and had two of the native-Bengali annotators who didn't participate in the annotation process, re-score them blind under the same rubric. They were compensated at standard per-sample rates. The
judge agreed with the annotator-majority score within
$\pm 1$ on 94.5\% of pairs and exactly on 71.0\%, with
a Spearman rank correlation of $\rho = 0.89$ between
judge and annotator-majority scores. This is
consistent with the agreement levels reported for
LLM-as-a-Judge on text-generation tasks and supports
the use of the score as a complementary metric, though
not as a replacement for CER/WER on this benchmark.

\paragraph{Limitations.}
The judge is itself a frontier VLM and may share blind spots with the 
systems it scores; in particular, it may be lenient toward outputs 
that drift into plausible but incorrect Bengali, since fluency is a 
strong prior for any large language model. A second, distinct concern 
is in-family bias: prior work has shown that LLM judges systematically 
favor outputs from their own family 
\citep{panickssery2024llm,wataoka2024selfpref,spiliopoulou2025playfavorites,li2025preference}. 
We address this for the two Gemini-family systems by re-scoring them 
with Claude Sonnet 4.6 as an out-of-family judge and using those 
scores in the main leaderboard, but we were unable to apply 
out-of-family re-scoring to the remaining thirteen models under our 
budget, so residual in-family inflation cannot be excluded for any 
non-Gemini cell. We therefore report judge scores alongside, not in 
place of, the edit-distance metrics, and treat large disagreements 
between the two (e.g., a model with moderate CER but a low judge 
score) as a diagnostic signal of semantic hallucination rather than 
as a ranking authority.



\section{Experimental Setup}
\label{app:setup}

\subsection{Evaluated Models}
\label{app:model_list}
Our evaluation covers fifteen VLMs across closed-source and open-source families (Tab.~\ref{tab:models}), together with three conventional OCR baselines (EasyOCR\rev{~\citep{jaidedai2020easyocr}}, Tesseract\rev{~\citep{smith2007tesseract}}, Surya\rev{~\citep{paruchuri2024surya}}). The fifteen VLMs are run across all three prompting conditions. Closed-source models are queried via their respective provider APIs with temperature $=0$ and a maximum of 150 completion tokens; images are submitted as base64-encoded data URLs. Open-source models are served locally in \texttt{bfloat16} with greedy decoding and the same 150-token cap. The fine-tuned and zero-shot evaluations share the same held-out test set. \rev{The OCR baselines are run without image preprocessing; versions, language settings, and detection and recognition components are listed in Tab.~\ref{tab:ocr_config}. Closed-source models were accessed under their providers' API terms. Open-weight models were used under their licenses (Qwen3.5 and InternVL3.5: Apache-2.0; Gemma~3: Gemma Terms of Use; Llama~4: Llama~4 Community License), as were the OCR tools (EasyOCR, Tesseract, and \texttt{pytesseract}: Apache-2.0; Surya: GPL-3.0 code with separately licensed weights that permit research use). All use was non-commercial research, consistent with these terms.}

\begin{table}[h]
\scriptsize
\centering
\setlength{\tabcolsep}{2.5pt}
\resizebox{\columnwidth}{!}{\begin{tabular}{@{}lllll@{}}
\toprule
\textbf{System} & \textbf{Version} & \textbf{Lang} & \textbf{Detection} & \textbf{Recognition} \\
\midrule
Tesseract & \texttt{pytesseract} 0.3.13 & \texttt{ben} & Tesseract & Tesseract \texttt{ben} \\
EasyOCR & 1.7.2 & \texttt{bn} & CRAFT~\citep{baek2019craft} & EasyOCR \texttt{bn} \\
Surya & 0.14.7 & \texttt{bn} & Surya & Surya v1 \\
\bottomrule
\end{tabular}}
\caption{\rev{Configuration of the conventional OCR baselines. No image preprocessing is applied.}}
\label{tab:ocr_config}
\end{table}

\subsection{Fine-Tuning Details}
\label{app:training}
We fine-tune the six open-source models listed in Tab.~\ref{tab:leaderboard}: InternVL3.5-2B, InternVL3.5-4B, InternVL3.5-8B, Gemma-3-4B-it, Gemma-3-12B-it, and Qwen3.5-9B. Closed-source models are excluded since their weights are not publicly accessible. \rev{The larger open models (Qwen3.5-27B, Qwen3.5-35B-A3B, and Llama~4 Maverick) were not fine-tuned within our compute budget.} LoRA adapters are attached to the attention projections (\texttt{q\_proj}, \texttt{k\_proj}, \texttt{v\_proj}, \texttt{o\_proj}) and MLP projections (\texttt{gate\_proj}, \texttt{up\_proj}, \texttt{down\_proj}) of the \emph{language-model component only}; the vision encoder and any vision-language adapter remain frozen. Base weights are loaded in \texttt{bfloat16} without quantization. We use the SFTTrainer from \texttt{trl}~\citep{vonwerra2020trl} with a custom collator that masks the user-prompt tokens so loss is computed only over assistant tokens.

We use a 50/10/40 train/validation/test split, with the same test split used for the zero-shot evaluation. Training runs for up to 10 epochs with early stopping on validation loss (patience $= 2$). The effective batch size is 16 (per-device 2, gradient accumulation 8), with AdamW, a peak learning rate of $2\times 10^{-4}$, cosine decay, and 10 warmup steps. All runs use a single seed (\texttt{seed} $= 3407$) on one NVIDIA A100 GPU. Full hyperparameters are listed in Tab.~\ref{tab:hyperparams}. 

\paragraph{\rev{EasyOCR fine-tuning.}}
\rev{To test whether adaptation changes a conventional recognizer's error profile, we fine-tune EasyOCR's CRNN recognizer~\citep{shi2017crnn} on the same 1{,}268-image training split, leaving the CRAFT detector unchanged, and evaluate it on the same test split with the identical scoring pipeline and error classifier (Appendix~\ref{app:classifier}).}

\begin{table}[h]
\scriptsize
\centering
\begin{tabular}{ll}
\toprule
\textbf{Hyperparameter} & \textbf{Value} \\
\midrule
LoRA rank $r$           & 16 \\
LoRA alpha $\alpha$     & 32 \\
LoRA dropout            & 0.05 \\
LoRA target modules     & $q, k, v, o$; \textit{up, down, gate} \\
Max epochs              & 10 \\
Early stopping patience & 2 \\
Batch size (per device) & 2 \\
Gradient accumulation   & 8 (eff.\ batch $= 16$) \\
Optimizer               & AdamW \\
Learning rate           & $2 \times 10^{-4}$ \\
LR schedule             & Cosine decay \\
Warmup steps            & 10 \\
Max sequence length     & 512 tokens \\
Precision               & \texttt{bfloat16} \\
GPU                     & NVIDIA A100 \\
Train / val / test      & 50 / 10 / 40 \\
\bottomrule
\end{tabular}
\caption{Hyperparameters used for LoRA fine-tuning across all models.}
\label{tab:hyperparams}
\end{table}

\begin{table}[h]
\scriptsize
\centering
\setlength{\tabcolsep}{3pt}
\begin{tabular}{@{}p{4.1cm}lc@{}}
\toprule
\textbf{Model} & \textbf{Provider} & \textbf{Access} \\
\midrule

\multicolumn{3}{@{}l}{\textit{Closed-source}} \\
Gemini 2.5 Pro~\citep{gemini2_5_2025}        & Google    & API \\
Gemini 2.5 Flash~\citep{gemini2_5_2025}      & Google    & API \\
Claude Sonnet 4.6~\citep{claude_sonnet_4_6_2026} & Anthropic & API \\
Claude Haiku 4.5~\citep{claude_haiku_4_5_2025}   & Anthropic & API \\
GPT-5.4-mini~\citep{gpt5_4_mini_nano_2026}   & OpenAI    & API \\
GPT-5.4-nano~\citep{gpt5_4_mini_nano_2026}   & OpenAI    & API \\

\midrule
\multicolumn{3}{@{}l}{\textit{Open-source}} \\
Qwen3.5-9B~\citep{qwen3_5_2026}              & Alibaba   & Open \\
Qwen3.5-27B~\citep{qwen3_5_2026}             & Alibaba   & Open \\
Qwen3.5-35B-A3B~\citep{qwen3_5_2026}         & Alibaba   & Open \\
Gemma-3-4B-it~\citep{Kamath2025Gemma3T}      & Google    & Open \\
Gemma-3-12B-it~\citep{Kamath2025Gemma3T}     & Google    & Open \\
InternVL3.5-2B~\citep{internvl3_5_2025}      & OpenGVLab & Open \\
InternVL3.5-4B~\citep{internvl3_5_2025}      & OpenGVLab & Open \\
InternVL3.5-8B~\citep{internvl3_5_2025}      & OpenGVLab & Open \\
Llama 4 Maverick~\citep{llama4_2025}         & Meta      & Open \\

\bottomrule
\end{tabular}
\caption{Vision-language models evaluated in our benchmark.}
\label{tab:models}
\end{table}

\section{Error-Classification Protocol}
\label{app:classifier}

This section specifies how each (ground-truth, prediction) pair is assigned to exactly one of the fifteen categories in Tab.~\ref{tab:error_taxonomy_defs}. No manual or LLM labeling enters at any stage; assignment is performed by a single deterministic, rule-based classifier, released with the code, and applied identically to all four VLM families and three conventional OCR baselines\rev{, including EasyOCR after fine-tuning}. Image-level events (script contamination, semantic hallucination, truncation, spurious extra content, numeric-only error) are counted once per image; word-level events come from images that produced predominantly correct Bengali but with localized errors, with one event per misaligned word.

\paragraph{Taxonomy construction.}
The taxonomy was developed inductively from a pilot error analysis of two representative systems (Gemini~2.5 Flash and Qwen3.5-9B) and cross-checked against prior Indic-OCR error typologies \citep{nayef2019icdar, gunna2022transferlearning}. We retained the conventional orthographic categories those typologies emphasize (matra substitution, consonant homophones, conjunct/hasanta errors, nasal diacritics, and compound orthographic edits) and added five categories specific to generative transcription that classical OCR typologies do not cover: script contamination, semantic hallucination, spurious extra content, truncation/no output, and numeric/temporal substitution. The category set was frozen before the full classification pass.

\paragraph{Determinism and cross-model consistency.}
The classifier receives only the pair (ground truth, prediction); it has no access to the source image, the prompt, or the identity of the model that produced the prediction. It therefore cannot assign the same (GT, prediction) input to different categories for different models, so cross-model consistency is guaranteed by construction rather than enforced by review. Given a fixed input the output is fully reproducible: there is no sampling, no data-fit threshold, and no human adjudication step. \rev{Because it sees only strings, the classifier characterizes the form of an error, not its origin within a model: category names such as \emph{visual grapheme misread} denote edit patterns consistent with visual mis-recognition, not verified causes.}

\paragraph{Stage 1: image-level cascade.}
Each prediction is first tested for whole-output failure modes in a fixed priority order, before any word-level analysis. In order: (i) \emph{truncation / no output} (empty string, refusal, or output length below 40\% of the ground-truth length); (ii) \emph{script contamination} (the dominant Unicode block of the output is Devanagari, Arabic, or Latin despite a Bengali ground truth); (iii) \emph{semantic hallucination} (the output is well-formed Bengali but shares under 25\% of its characters with the ground truth); (iv) \emph{spurious extra content} (the full ground-truth string is recovered but accompanied by unsolicited additional tokens, such as an appended translation or filler); and (v) \emph{numeric / temporal error} (all non-numeric content matches after applying a numeric mask, and only Bengali digits differ). The first rule that fires assigns the category, and the event is counted once for the image. If none fires, the image is judged to contain predominantly correct Bengali with localized errors and is passed to Stage~2.

\paragraph{Stage 2: word-level operation counting.}
For images reaching Stage~2, the ground-truth and predicted strings are aligned word-by-word via a Levenshtein alignment. Each misaligned word is then aligned at the character level, and its edit operations (insertions, deletions, substitutions) are partitioned by the Unicode category of the characters involved: base consonants, dependent vowel signs (matra), hasanta/virama, nasal diacritics (chandrabindu, anusvara, visarga), and digits. One event is emitted per misaligned word.

\paragraph{Hierarchical resolution to a single class.}
Each word-level event is resolved to exactly one category by a total, mutually exclusive decision hierarchy over the edit statistics, using the per-class thresholds listed in Tab.~\ref{tab:error_taxonomy_defs}. Word-magnitude classes are tested first: a word is \emph{lexical word substitution} when character overlap falls below 40\% or the edit distance reaches 60\% of the word length; \emph{multi-character word corruption} when there are at least three grapheme edits with over 40\% character overlap retained; and \emph{visual grapheme misread} when a single visually-similar but linguistically-unrelated substitution occurs within an edit distance of two. If the edits are instead confined to a single orthographic Unicode category, the event resolves to the corresponding orthographic class: matra-only edits to \emph{vowel-sign substitution}; hasanta or conjunct-ligature edits to \emph{conjunct/hasanta error}; substitutions within a fixed homophone-class lookup (\banglafont{ন/ণ} \textit{(na/ṇa)}, \banglafont{স/শ/ষ} \textit{(sa/śa/ṣa)}, \banglafont{ব/ভ} \textit{(ba/bha)}, \banglafont{ত/থ} \textit{(ta/tha)}, and similar) to \emph{consonant homophone}; nasal-diacritic-only edits to \emph{nasal diacritic error}; and edits spanning two or more orthographic categories (with no unrelated consonant substitution) to \emph{compound orthographic}. Pairs that differ only after NFC normalization or in punctuation/quote glyphs resolve to \emph{unicode normalization}. Because the hierarchy is exhaustive and its branches are disjoint, every event maps to exactly one of the fifteen categories.

\paragraph{Counting convention.}
Image-level events (script contamination, semantic hallucination, truncation/no output, spurious extra content, and numeric/temporal error) are counted once per image; word-level events are counted once per misaligned word. The per-model percentages in Tab.~\ref{tab:error_taxonomy} are the share of each category among all events attributed to a given system, and the visual-to-orthographic ratios reported in \S~\ref{sec:error} are computed from these shares. \rev{Tab.~\ref{tab:error_taxonomy} groups the categories into word-level and image-level panels with per-level subtotals and also reports the visual-consistent and orthographic categories as shares of word-level events, which is the basis of the percentages in \S\ref{sec:error}. Because both groups are word-level, their ratio does not depend on how image-level events are counted.}

\vspace{4em}
\section{Error Analysis}
\label{app:error}
Three examples of each of the error categories specified in Tab.~\ref{tab:error_taxonomy} can be found in Tab.~\ref{tab:taxonomy_examples_a} and Tab.~\ref{tab:taxonomy_examples_b}.

\section{Error Taxonomy}
The definitions of each of the error categories specified in Tab.~\ref{tab:error_taxonomy} can be found in Tab.~\ref{tab:error_taxonomy_defs}.
\begin{table*}[t]
\centering
\small
\setlength{\tabcolsep}{6pt}
\renewcommand{\arraystretch}{1.2}
\begin{tabular}{p{3.3cm} p{\dimexpr\textwidth-3.3cm-4\tabcolsep\relax}}
\toprule
\textbf{Error type} & \textbf{Description} \\
\midrule
Lexical word substitution &
Whole word replaced by an unrelated lexical item (char overlap $<\!40\%$ or edit distance $\geq\!60\%$ of length). \del{Visually-driven recognition failure at the word level} \rev{Pattern consistent with visual mis-recognition at the word level}, e.g.\ \banglafont{ফ্ল্যাটটি}(\textit{phlyāṭṭi})$\to$\banglafont{ফ্যাক্টরি}(\textit{phyākṭari}), \banglafont{কপাট}(\textit{kapāṭa})$\to$\banglafont{লোহ}(\textit{loha}), \banglafont{বৈষম্য}(\textit{baiṣamya})$\to$\banglafont{লোন} (\textit{lona}). \\

Visual grapheme misread &
Single-character substitution between visually-similar but linguistically-unrelated graphemes (edit distance $\leq\!2$ over a word). \del{Reflects feature-level vision-encoder error} \rev{Pattern consistent with a visual confusion between graphemes}, e.g.\ \banglafont{মেধা}  (\textit{medhā})$\to$\banglafont{মেবা}(\textit{mebā}), \banglafont{চুইঝাল}(\textit{cuijhāla})$\to$\banglafont{চুইমাল} (\textit{cuimāla}). \\

Word-boundary segmentation &
Adjacent words concatenated (\banglafont{এনা} (\textit{enā})$\to$\banglafont{এন} (\textit{ena})~\banglafont{এ} (\textit{e})) or a single token spuriously split (\banglafont{পাটকল} (\textit{pāṭakala})$\to$\banglafont{পটল কল} (\textit{paṭala kala})); also covers whitespace-only normalization. Inflates WER but leaves the character sequence largely intact. \\

Multi-character word corruption &
Word recognizable in shape but with $\geq\!3$ grapheme edits and $>\!40\%$ character overlap (e.g.\ \banglafont{পাঞ্জেরী} (\textit{pāñjerī})$\to$\banglafont{পান্তরী} (\textit{pāntarī}), \banglafont{স্বাধীনতা} (\textit{svādhīnatā})$\to$\banglafont{সাবধানতা} (\textit{sābadhanatā}). Partial \del{visual} recovery without lexical replacement. \\

Vowel-sign (matra) substitution &
Only dependent vowel signs differ between GT and prediction (\banglafont{ি/ী} (i/ī), \banglafont{ু/ূ} (u/ū), \banglafont{ে/ৈ} (e/ai), \banglafont{ো/ৌ} (o/au)). Base consonants and conjuncts are preserved. \\

Script contamination &
Output drifts into Devanagari, Arabic, or dominant Latin transliteration despite Bengali input. Distinct cross-lingual failure mode. \\

Unicode normalization &
Strings differ only after NFC normalization, or differ in punctuation/quote glyphs (\textit{e.g.} \texttt{"} vs \texttt{“}, \texttt{-} vs \texttt{--}). Penalized by CER/WER but visually identical. \\

Consonant homophone &
Substitution within a known orthographic-homophone class (\banglafont{ন/ণ} (\textit{na/ṇa}), \banglafont{স/শ/ষ} (\textit{sa/śa/ṣa}), \banglafont{ব/ভ}(\textit{ba/bha}), \banglafont{ত/থ} (\textit{ta/tha}), etc.). Pure spelling-knowledge failure with the right phonetic target. \\

Conjunct / hasanta error &
Hasanta (\banglafont{◌্}) dropped or spuriously inserted, conjunct ligature mis-segmented (\banglafont{স্ব} (\textit{sva})$\to$\banglafont{সব} (\textit{saba}), \banglafont{ক্ষ} (\textit{kṣa})$\to$\banglafont{খ} (\textit{kha}), \banglafont{ব্র} (\textit{bra})$\to$\banglafont{বর} (\textit{bara})). Leaves the base consonant identities ambiguous. \\

Compound orthographic &
Multiple orthographic-type errors (matra + nasal + homophone) co-occur in a single word, with no unrelated consonant substitution. \\

Semantic hallucination &
Whole-image output is \del{fluent} \rev{well-formed} Bengali but shares $<\!25\%$ characters with GT (\rev{consistent with a} language-prior takeover, e.g.\ \banglafont{ছাত্র ইউনিয়ন} (\textit{chātra iuniyana})$\to$\banglafont{শুভ জন্মদিন} (\textit{śubha janmadina})). \\

Numeric / temporal error &
All non-numeric text correct; only Bengali numerals are wrong (\banglafont{১৪৩৩} (\textit{1433})$\to$\banglafont{১৪৩০} (\textit{1430}), \banglafont{২০২৫} (\textit{2025})$\to$\banglafont{২০২৩} (\textit{2023})). \\

Nasal diacritic error &
Chandrabindu (\banglafont{ঁ}), anusvara (\banglafont{ং}), or visarga (\banglafont{ঃ}) confused or dropped, no other change. \\

Truncation / no output &
Empty output, refusal, or length below 40\% of ground truth. \\

Spurious extra content &
GT text correctly present plus unsolicited translation or filler. \\
\bottomrule
\end{tabular}
\caption{Definitions of the fifteen error categories in the \textsc{BanglaWild} taxonomy (see Table~\ref{tab:error_taxonomy} for per-model rates). Bengali examples are shown with ALA-LC romanization. \rev{Descriptions characterize string-level edit patterns; they are consistent with, but do not establish, the causes they mention.}}
\label{tab:error_taxonomy_defs}
\end{table*}
\begin{table*}[t]
\centering
\footnotesize
\setlength{\tabcolsep}{4pt}
\renewcommand{\arraystretch}{1.1}
\resizebox{\textwidth}{!}{
\begin{tabular}{@{}p{3.8cm} p{0.7cm} p{4.6cm} p{4.6cm}@{}}
\toprule
\textbf{Error type} & \textbf{M} & \textbf{Ground truth} & \textbf{Model prediction} \\
\midrule

\multirow{3}{=}{\textbf{Lexical word substitution} \newline {\footnotesize\itshape Whole word replaced by an unrelated lexical item.}}
& G & \banglafont{কল্যাণ রাষ্ট্র} \underline{\banglafont{চাই}} \newline \textit{kalyāṇa rāṣṭra \underline{cāi}}
  & \underline{\banglafont{কল্যাণপুর}} \newline \textit{\underline{kalyāṇapura}} \\
& G & \banglafont{সব শিশুরা} \underline{\banglafont{হাদি}} \banglafont{হবে!} \newline \textit{saba śiśurā \underline{hādi} habe!}
  & \banglafont{সব শ্রমিক} \underline{\banglafont{এক}} \banglafont{হবে!} \newline \textit{saba śramika \underline{eka} habe!} \\
& Q & \banglafont{অসাম্প্রদায়িক বাংলাদেশ} \newline \textit{asāmpradāyika bāṃlādeśa}
  & \underline{\banglafont{আমাদের}} \banglafont{বাংলাদেশ} \newline \textit{\underline{āmādera} bāṃlādeśa} \\

\midrule

\multirow{3}{=}{\textbf{Visual grapheme misread} \newline {\footnotesize\itshape Single-character substitution between similar graphemes.}}
& G & \underline{\banglafont{গদি}} \banglafont{ছাড়} \newline \textit{\underline{gadi} chāṛa}
  & \underline{\banglafont{সদি}} \banglafont{ছাড়} \newline \textit{\underline{sadi} chāṛa} \\
& G & \banglafont{জান দেবো} \underline{\banglafont{জুলাই}} \banglafont{দেবোনা} \newline \textit{jāna debō \underline{julāi} debonā}
  & \banglafont{জান দেবো} \underline{\banglafont{জুলুম}} \banglafont{দেবো না} \newline \textit{jāna debō \underline{juluma} debō nā} \\
& Q & \banglafont{ধর্ম যার যার দেশটা} \underline{\banglafont{সবার}} \newline \textit{dharma yāra yāra deśaṭā \underline{sabāra}}
  & \banglafont{ধর্ম যার যার দেশটা} \underline{\banglafont{আমার}} \newline \textit{dharma yāra yāra deśaṭā \underline{āmāra}} \\

\midrule

\multirow{3}{=}{\textbf{Word-boundary segmentation} \newline {\footnotesize\itshape Adjacent words concatenated or a single token split.}}
& G & \underline{\banglafont{সমঅধিকার-সমমর্যাদা}} \banglafont{নিশ্চিত কর} \newline \textit{\underline{samaadhikāra-samamaryādā} niścita kara}
  & \underline{\banglafont{সম অধিকার- সম মর্যাদা}} \banglafont{নিশ্চিত কর} \newline \textit{\underline{sama adhikāra- sama maryādā} niścita kara} \\
& G & \underline{\banglafont{অটোপরিবহন}} \newline \textit{\underline{aṭōparibahana}}
  & \underline{\banglafont{অটো পরিবহন}} \newline \textit{\underline{aṭō paribahana}} \\
& G & \banglafont{রাষ্ট্রীয়}\underline{\banglafont{ও}} \banglafont{দলীয় সন্ত্রাস মুক্ত}\dots \newline \textit{rāṣṭrīẏa\underline{o} dalīẏa santrāsa mukta\dots}
  & \banglafont{রাষ্ট্রীয়} \underline{\banglafont{ও}} \banglafont{দলীয় সন্ত্রাস মুক্ত}\dots \newline \textit{rāṣṭrīẏa \underline{o} dalīẏa santrāsa mukta\dots} \\

\midrule

\multirow{3}{=}{\textbf{Multi-character word corruption} \newline {\footnotesize\itshape Word recognizable but with $\geq\!3$ grapheme edits.}}
& G & \banglafont{কালচারাল} \underline{\banglafont{ফ্যাসিজম}} \banglafont{নিপাত যাক} \newline \textit{kālacārāla \underline{phyāsijama} nipāta yāka}
  & \banglafont{কালচারাল} \underline{\banglafont{ফ্যাসিবাদ}} \banglafont{নিপাত যাক} \newline \textit{kālacārāla \underline{phyāsibāda} nipāta yāka} \\
& G & \banglafont{দাবি\dots চাই} \underline{\banglafont{নির্দলীয়}} \banglafont{সরকার} \newline \textit{dābi\dots cāi \underline{nirdalaīẏa} sarakāra}
  & \banglafont{দাবি\dots চাই} \underline{\banglafont{নিরালা}} \banglafont{সরকার} \newline \textit{dābi\dots cāi \underline{nirālā} sarakāra} \\
& Q & \underline{\banglafont{স্বৈরাচার}} \banglafont{মুক্ত বাংলাদেশ} \newline \textit{\underline{sbairācāra} mukta bāṃlādeśa}
  & \underline{\banglafont{মেরচার}} \banglafont{মুক্ত বাংলাদেশ} \newline \textit{\underline{meracāra} mukta bāṃlādeśa} \\

\midrule

\multirow{3}{=}{\textbf{Vowel-sign substitution} \newline {\footnotesize\itshape Only dependent vowel signs differ.}}
& G & \underline{\banglafont{শহিদ}} \banglafont{মিনার আবাসিক এলাকা} \newline \textit{\underline{śahida} mināra ābāsika elākā}
  & \underline{\banglafont{শহীদ}} \banglafont{মিনার আবাসিক এলাকা} \newline \textit{\underline{śahīda} mināra ābāsika elākā} \\
& G & \banglafont{লাখো\dots কেনা} \underline{\banglafont{দেশটা}} \banglafont{কারো} \newline \textit{lākho\dots kenā \underline{deśaṭā} kāro}
  & \banglafont{লাখো\dots কেনা} \underline{\banglafont{দেশটি}} \banglafont{কারো} \newline \textit{lākho\dots kenā \underline{deśaṭi} kāro} \\
& Q & \underline{\banglafont{পাল্টা}} \banglafont{আঘাত} \newline \textit{\underline{pālṭā} āghāta}
  & \underline{\banglafont{পল্টা}} \banglafont{আঘাত} \newline \textit{\underline{palṭā} āghāta} \\

\midrule

\multirow{3}{=}{\textbf{Script contamination} \newline {\footnotesize\itshape Output drifts into Devanagari, Arabic, or Latin.}}
& G & \banglafont{যুদ্ধ ৯ শান্তি} \newline \textit{yuddha o śānti}
  & \textsc{war vs peace} \\
& G & \banglafont{রাখতে} \newline \textit{rākhate}
  & \textsc{sempre} \\
& Q & \banglafont{হাল ছেড়োনা\dots কন্ঠ ছাড়ো জোরে} \newline \textit{hāla cheṛonā\dots kanṭha chāṛo jore}
  & \foreignlanguage{hindi}{तुमने} \\

\midrule

\multirow{3}{=}{\textbf{Unicode normalization} \newline {\footnotesize\itshape Strings differ only after NFC or in punctuation glyphs.}}
& G & \banglafont{ধর্মীয় হিংসা নয়\dots সাম্প্রদায়িক}\dots \newline \textit{dharmīẏa hiṃsā naya\dots sāmpradāyika\dots}
  & \banglafont{ধর্মীয় হিংসা নয়\dots সাম্প্রদায়িক}\dots \newline \textit{dharmīẏa hiṃsā naya\dots sāmpradāyika\dots} \\
& G & \banglafont{নাটক কম} \underline{\banglafont{করো}} \banglafont{পিও!} \newline \textit{nāṭaka kama \underline{karo} pio!}
  & \banglafont{নাটক কম} \underline{\banglafont{করো}} \banglafont{পিও!} \newline \textit{nāṭaka kama \underline{karo} pio!} {\scriptsize(NFC-only)} \\
& Q & \banglafont{ধর্মীয় হিংসা} \underline{\banglafont{নয়\dots}} \banglafont{সাম্প্রদায়িক}\dots \newline \textit{dharmīẏa hiṃsā \underline{naya\dots} sāmpradāyika\dots}
  & \banglafont{ধর্মীয় হিংসা} \underline{\banglafont{নয়\dots}} \banglafont{সাম্প্রদায়িক}\dots \newline \textit{dharmīẏa hiṃsā \underline{naya\dots} sāmpradāyika\dots} \\

\bottomrule
\end{tabular}}
\caption{Three real examples per error category (part 1 of 2), drawn from the model outputs of \textbf{G}=Gemini-2.5-Flash and \textbf{Q}=Qwen3.5-9B. For word-level categories the affected word is \underline{underlined}; for image-level categories the full transcription is shown. Each cell shows Bengali script (top) and ALA-LC romanization (bottom, \textit{italic}); differing characters are \underline{underlined}. Romanization follows ALA-LC 2017~\cite{alalc2017bengali}. \textit{Continued in Table~\ref{tab:taxonomy_examples_b}.}}
\label{tab:taxonomy_examples_a}
\end{table*}

\begin{table*}[t]
\centering
\footnotesize
\setlength{\tabcolsep}{4pt}
\renewcommand{\arraystretch}{1.1}
\resizebox{\textwidth}{!}{
\begin{tabular}{@{}p{3.8cm} p{0.7cm} p{4.6cm} p{4.6cm}@{}}
\toprule
\textbf{Error type} & \textbf{M} & \textbf{Ground truth} & \textbf{Model prediction} \\
\midrule

\multirow{3}{=}{\textbf{Consonant homophone} \newline {\footnotesize\itshape Within a known homophone class (\banglafont{ন/ণ}, \banglafont{স/শ/ষ}\dots).}}
& G & \underline{\banglafont{সাধারন}} \banglafont{সম্পাদক} \newline \textit{\underline{sādhārana} sampādaka}
  & \underline{\banglafont{সাধারণ}} \banglafont{সম্পাদক} \newline \textit{\underline{sādhāraṇa} sampādaka} \\
& G & \banglafont{রক্ত গরম মাথা} \underline{\banglafont{ঠান্ডা}}! \newline \textit{rakta garama māthā \underline{ṭhānḍā}!}
  & \banglafont{রক্ত গরম মাথা} \underline{\banglafont{ঠাণ্ডা}}! \newline \textit{rakta garama māthā \underline{ṭhāṇḍā}!} \\
& Q & \underline{\banglafont{স্লোগান}} \banglafont{তোমার বাংলাদেশ সবার} \newline \textit{\underline{slōgāna} tomāra bāṃlādeśa sabāra}
  & \underline{\banglafont{শ্লোগান}} \banglafont{তোমার বাংলাদেশ আমার} \newline \textit{\underline{ślōgāna} tomāra bāṃlādeśa āmāra} \\

\midrule

\multirow{3}{=}{\textbf{Conjunct / hasanta error} \newline {\footnotesize\itshape Hasanta dropped or conjunct mis-segmented.}}
& G & \underline{\banglafont{আল্লাহ্}} \banglafont{ভরসা} \newline \textit{\underline{āllāh\textsuperscript{◌্}} bharasā}
  & \underline{\banglafont{আল্লাহ}} \banglafont{ভরসা} \newline \textit{\underline{āllāha} bharasā} \\
& G & \underline{\banglafont{স্বাধীনতা}} \newline \textit{\underline{sbādhīnatā}}
  & \underline{\banglafont{সাবধানতা}} \newline \textit{\underline{sābadhānatā}} \\
& Q & \banglafont{আমাদের} \underline{\banglafont{স্বাধীন}} \banglafont{দেশে\dots} \newline \textit{āmādera \underline{sbādhīna} deśe\dots}
  & \banglafont{আমাদের} \underline{\banglafont{সুবিধীন}} \banglafont{দেশে\dots} \newline \textit{āmādera \underline{subidhīna} deśe\dots} \\

\midrule

\multirow{3}{=}{\textbf{Compound orthographic} \newline {\footnotesize\itshape Multiple orthographic-type errors in one word.}}
& G & \underline{\banglafont{তারূন্যের}} \banglafont{প্রথম ভোট} \newline \textit{\underline{tārūnyera} prathama bhoṭa}
  & \underline{\banglafont{তারুণ্যের}} \banglafont{প্রথম ভোট} \newline \textit{\underline{tāruṇyera} prathama bhoṭa} \\
& G & \banglafont{মডার্ণ} \underline{\banglafont{ষ্টেশনারী}} \newline \textit{maḍārṇa \underline{ṣṭeśanārī}}
  & \banglafont{মডার্ন} \underline{\banglafont{স্টেশনারি}} \newline \textit{maḍārna \underline{sṭeśanāri}} \\
& Q & \banglafont{২৪ এর\dots} \underline{\banglafont{তরূণ}} \banglafont{যেন\dots} \newline \textit{24 era\dots \underline{tarūṇa} yena\dots}
  & \banglafont{২৪ এর\dots} \underline{\banglafont{তরুন}} \banglafont{যেন\dots} \newline \textit{24 era\dots \underline{taruna} yena\dots} \\

\midrule

\multirow{3}{=}{\textbf{Semantic hallucination} \newline {\footnotesize\itshape Fluent Bengali, $<\!25\%$ char overlap with GT.}}
& G & \banglafont{৩৬শে জুলাই} \newline \textit{36śe julāi}
  & \banglafont{ভোট দিন} \newline \textit{bhoṭa dina} \\
& G & \banglafont{ওয়েবসাইট লাগবে?} \newline \textit{oyebosāiṭa lāgabe?}
  & \banglafont{ধূমপান?} \newline \textit{dhūmapāna?} \\
& Q & \banglafont{স্বাদ আর বিশুদ্ধতায় সেরা} \newline \textit{sbāda āra biśuddhatāẏa serā}
  & \banglafont{মাদার} \newline \textit{mādāra} \\

\midrule

\multirow{3}{=}{\textbf{Numeric / temporal error} \newline {\footnotesize\itshape Surrounding text correct; numerals wrong.}}
& G & \banglafont{শুভ নববর্ষ} \underline{\banglafont{১৪৩৩}} \newline \textit{śubha nababarṣa \underline{1433}}
  & \banglafont{শুভ নববর্ষ} \underline{\banglafont{১৪৩০}} \newline \textit{śubha nababarṣa \underline{1430}} \\
& G & \banglafont{মাসব্যাপী\dots উৎসব} \underline{\banglafont{২০২৫}} \newline \textit{māsabyāpī\dots utsaba \underline{2025}}
  & \banglafont{মাসব্যাপী\dots উৎসব} \underline{\banglafont{২০২৩}} \newline \textit{māsabyāpī\dots utsaba \underline{2023}} \\
& Q & \banglafont{মেরামতের} \underline{\banglafont{৩১}} \banglafont{দফা} \newline \textit{merāmatera \underline{31} daphā}
  & \banglafont{মেরামতের} \underline{\banglafont{৬১}} \banglafont{দফা} \newline \textit{merāmatera \underline{61} daphā} \\

\midrule

\multirow{3}{=}{\textbf{Nasal diacritic error} \newline {\footnotesize\itshape Chandrabindu, anusvara, or visarga differs.}}
& G & \banglafont{ফুল} \underline{\banglafont{ছেড়া}} \banglafont{নিষেধ সুজন} \newline \textit{phula \underline{cheṛā} niṣedha sujana}
  & \banglafont{ফুল} \underline{\banglafont{ছেঁড়া}} \banglafont{নিষেধ সুজন} \newline \textit{phula \underline{chem̐ṛā} niṣedha sujana} \\
& G & \banglafont{বৈষম্যের} \underline{\banglafont{ঠাঁই}} \banglafont{নাই} \newline \textit{baiṣamyera \underline{ṭhām̐i} nāi}
  & \banglafont{বেগমোর} \underline{\banglafont{ঠাই}} \banglafont{নাই} \newline \textit{begamora \underline{ṭhāi} nāi} \\
& G & \banglafont{আড়ং ন্যাচারাল} \underline{\banglafont{চিনিগুড়া}} \banglafont{চাল} \newline \textit{āṛaṃ nyācārāla \underline{ciniguṛā} cāla}
  & \banglafont{আড়ং ন্যাচারাল} \underline{\banglafont{চিনিগুঁড়া}} \banglafont{চাল} \newline \textit{āṛaṃ nyācārāla \underline{cinigum̐ṛā} cāla} \\

\midrule

\multirow{3}{=}{\textbf{Truncation / no output} \newline {\footnotesize\itshape Empty refusal or length $<\!40\%$ of GT.}}
& G & \banglafont{বল বীর চির উন্নত মম শির} \newline \textit{bala bīra cira unnata mama śira}
  & \banglafont{জন্মভূমির} \newline \textit{janmabhūmira} \\
& Q & \banglafont{শহীদের রক্ত হবে নাকো ব্যর্থ।} \newline \textit{śahīdera rakta habe nāko byartha.}
  & \banglafont{বঙ্গবন্ধু} \newline \textit{baṃgabandhu} \\
& Q & \banglafont{চির উন্নত মম শির} \newline \textit{cira unnata mama śira}
  & \banglafont{চিরতরে} \newline \textit{ciratare} \\

\midrule

\multirow{3}{=}{\textbf{Spurious extra content} \newline {\footnotesize\itshape GT text correct plus unsolicited additions.}}
& G & \banglafont{কবির পরিবহন} \newline \textit{kabira paribahana}
  & \banglafont{কবির পরিবহন} \underline{\banglafont{আশরাফুল ও আলামিন}} \newline \textit{kabira paribahana \underline{āśarāphula o ālāmina}} \\
& Q & \banglafont{কাজী অফিস} \newline \textit{kājī aphisa}
  & \underline{\banglafont{৫.১১}} \banglafont{কাজী অফিস} \newline \textit{\underline{5.11} kājī aphisa} \\
& Q & \banglafont{আশরাফ আলী} \newline \textit{āśarāpha ālī}
  & \banglafont{আশরাফ আলী} \underline{\banglafont{দা কুশ্চা}} \newline \textit{āśarāpha ālī \underline{dā kuścā}} \\

\bottomrule
\end{tabular}}
\caption{Three real examples per error category (part 2 of 2), continued from Table~\ref{tab:taxonomy_examples_a}; notation as in part 1.}
\label{tab:taxonomy_examples_b}
\end{table*}

\begin{table*}[htbp]
\centering
\caption{%
  Model benchmarking results (CER~\%) on the \textsc{BanglaWild} test set across
  \textbf{Content} and \textbf{Context} attribute categories under three prompting strategies:
  \textbf{P1}~(Baseline English), \textbf{P2}~(Baseline Bengali), and
  \textbf{P3}~(Structured English).
  \colorbox{bestblue}{Cyan} highlights the lowest CER per column;
  \textbf{bold} marks the best-performing model per group.%
}
\label{tab:bangla_wild_results}
\renewcommand{\arraystretch}{0.88}

\setlength{\tabcolsep}{4pt}
\adjustbox{max width=0.95\textwidth}{%
\begin{tabular}{l ccc ccc ccc ccc ccc ccc}
\toprule

& \multicolumn{18}{c}{\textbf{Content}} \\
\cmidrule(lr){2-19}

& \multicolumn{3}{c}{\makecell{\textit{Advertisement}}}
& \multicolumn{3}{c}{\makecell{\textit{Colloquial}}}
& \multicolumn{3}{c}{\makecell{\textit{Creative}}}
& \multicolumn{3}{c}{\makecell{\textit{Instructional}}}
& \multicolumn{3}{c}{\makecell{\textit{Political}}}
& \multicolumn{3}{c}{\makecell{\textit{Religious}}} \\

\cmidrule(lr){2-4}\cmidrule(lr){5-7}\cmidrule(lr){8-10}
\cmidrule(lr){11-13}\cmidrule(lr){14-16}\cmidrule(lr){17-19}

\textbf{Model}
& P1 & P2 & P3  & P1 & P2 & P3  & P1 & P2 & P3
& P1 & P2 & P3  & P1 & P2 & P3  & P1 & P2 & P3 \\

\midrule

Claude Haiku 4.5
  &  68.80 &  68.12 &  62.37
  & 183.96 & 146.16 & 140.16
  & 100.68 &  93.62 & 102.46
  &  63.35 &  62.91 &  65.55
  & 120.96 & 102.08 & 136.42
  &  46.15 &  45.91 &  45.94 \\

Claude Sonnet 4.6
  & \cellcolor{bestblue}\textbf{14.88} & \cellcolor{bestblue}\textbf{12.33} & \cellcolor{bestblue}\textbf{12.44}
  & 30.26 & 34.13 & 27.40
  & 24.14 & 21.77 & 19.58
  & 17.52 & 11.85 & \cellcolor{bestblue}\textbf{10.81}
  & 41.07 & 25.28 & 27.05
  & 13.81 & 12.79 & 13.17 \\

GPT-5.4-mini
  & 42.31 & 43.79 & 43.55
  & 66.71 & 81.33 & 65.55
  & 63.65 & 67.69 & 62.30
  & 39.87 & 48.19 & 37.85
  & 76.29 & 78.55 & 71.49
  & 31.14 & 32.68 & 31.70 \\

GPT-5.4-nano
  & 208.31 & 168.39 & 110.27
  & 381.47 & 258.86 & 128.38
  & 273.31 & 207.62 & 121.74
  & 213.33 & 187.52 & 109.71
  & 338.54 & 246.69 & 127.00
  & 117.75 & 123.29 &  94.27 \\

Gemini 2.5 Flash
  & 12.42 & 12.32 & 12.32
  & \cellcolor{bestblue}\textbf{21.69} & 27.91 & \cellcolor{bestblue}\textbf{20.82}
  & \cellcolor{bestblue}\textbf{13.39} & \cellcolor{bestblue}\textbf{13.43} & \cellcolor{bestblue}\textbf{13.58}
  & \cellcolor{bestblue}\textbf{8.76}  & \cellcolor{bestblue}\textbf{8.89}  & 8.76
  & \cellcolor{bestblue}\textbf{17.42} & \cellcolor{bestblue}\textbf{21.01} & \cellcolor{bestblue}\textbf{17.70}
  & \cellcolor{bestblue}\textbf{11.94} & 14.26 & \cellcolor{bestblue}\textbf{11.02} \\

Gemini 2.5 Pro
  & 21.05 & 18.40 & 23.54
  & 27.11 & 23.59 & 23.98
  & 28.79 & 25.05 & 22.72
  & 27.83 & 17.71 & 21.84
  & 26.43 & 24.29 & 27.42
  & 24.04 & 21.09 & 22.28 \\

Gemma-3-12B-it
  &  70.07 &  97.92 &  47.92
  &  60.06 &  62.65 &  61.30
  &  63.90 &  51.76 &  46.97
  &  25.65 &  28.58 &  62.15
  &  57.35 &  67.14 &  63.88
  &  72.68 &  56.81 &  45.88 \\

Gemma-3-4B-it
  & 122.19 &  89.70 &  68.22
  &  83.34 &  88.13 &  77.88
  & 165.02 & 116.95 &  70.39
  &  43.33 &  44.85 &  43.49
  & 160.29 & 178.83 &  80.65
  & 142.01 & 146.18 &  78.88 \\

InternVL3.5-2B
  & 114.03 & 168.66 & 121.81
  & 158.72 & 227.48 & 132.95
  & 147.21 & 177.66 & 142.14
  & 136.02 & 178.41 & 144.69
  & 153.90 & 194.03 & 155.85
  & 112.81 & 144.18 & 131.49 \\

InternVL3.5-4B
  & 114.15 & 156.77 & 112.18
  & 171.74 & 233.29 & 110.87
  & 112.74 & 148.57 & 116.33
  & 117.14 & 148.06 & 120.74
  & 117.03 & 159.41 & 116.29
  & 111.53 & 130.02 & 120.70 \\

InternVL3.5-8B
  & 108.46 & 116.59 & 103.10
  & 113.78 & 141.16 & 106.58
  & 123.25 & 116.43 & 101.60
  & 100.69 & 101.04 & 112.01
  & 135.60 & 122.04 & 105.51
  & 104.73 & 103.91 & 103.35 \\

Llama 4 Maverick
  & 32.89 & 36.31 & 36.08
  & 64.97 & 56.43 & 62.02
  & 38.14 & 39.48 & 38.97
  & 23.23 & 30.32 & 22.94
  & 45.21 & 44.76 & 46.36
  & 23.65 & 23.79 & 20.70 \\

Qwen3.5-9B
  & 16.13 & 15.80 & 15.69
  & 31.27 & 30.01 & 29.05
  & 22.21 & 22.34 & 23.02
  & 13.90 & 13.80 & 13.46
  & 27.71 & 28.90 & 26.95
  & 14.03 & 14.50 & 13.99 \\

Qwen3.5-27B
  & 18.19 & 18.02 & 18.54
  & 39.10 & 34.60 & 37.43
  & 26.09 & 27.10 & 25.62
  & 15.85 & 15.00 & 14.31
  & 33.15 & 32.32 & 33.00
  & 15.87 & 15.77 & 16.13 \\

Qwen3.5-35B-A3B
  & 17.27 & 15.92 & 16.68
  & 34.72 & 34.37 & 31.46
  & 21.79 & 21.54 & 21.57
  & 13.90 & 13.74 & 12.83
  & 28.33 & 27.21 & 27.81
  & 14.97 & \cellcolor{bestblue}\textbf{13.61} & \cellcolor{bestblue}\textbf{14.20} \\

\bottomrule
\end{tabular}%
}

\medskip

\setlength{\tabcolsep}{6.5pt}
\adjustbox{max width=0.95\textwidth}{%
\begin{tabular}{l ccc ccc ccc ccc ccc}
\toprule

& \multicolumn{15}{c}{\textbf{Context}} \\
\cmidrule(lr){2-16}

& \multicolumn{3}{c}{\makecell{\textit{Packaging}}}
& \multicolumn{3}{c}{\makecell{\textit{Printed Material}}}
& \multicolumn{3}{c}{\makecell{\textit{Signboard}}}
& \multicolumn{3}{c}{\makecell{\textit{Vehicle}}}
& \multicolumn{3}{c}{\makecell{\textit{Wall}}} \\

\cmidrule(lr){2-4}\cmidrule(lr){5-7}\cmidrule(lr){8-10}
\cmidrule(lr){11-13}\cmidrule(lr){14-16}

\textbf{Model}
& P1 & P2 & P3  & P1 & P2 & P3  & P1 & P2 & P3
& P1 & P2 & P3  & P1 & P2 & P3 \\

\midrule

Claude Haiku 4.5
  &  72.92 &  67.97 &  67.15
  &  51.95 &  47.45 &  52.55
  &  79.08 &  79.41 &  63.62
  & 126.45 & 120.66 & 110.74
  &  96.83 &  85.66 &  99.22 \\

Claude Sonnet 4.6
  & 16.35 & 12.73 & 14.44
  & 11.01 &  9.79 &  9.21
  & 18.80 & 12.47 & 12.77
  & 27.93 & 25.97 & 24.31
  & 26.19 & 20.77 & 20.04 \\

GPT-5.4-mini
  & 50.51 & 50.49 & 52.00
  & 33.30 & 34.58 & 31.25
  & 47.70 & 48.02 & 45.15
  & 56.95 & 63.44 & 58.60
  & 57.86 & 62.20 & 56.87 \\

GPT-5.4-nano
  & 198.38 & 128.76 & 109.40
  & 190.13 & 159.39 & 107.46
  & 236.35 & 178.03 & 108.37
  & 202.15 & 184.12 & 121.14
  & 269.10 & 208.90 & 117.09 \\

Gemini 2.5 Flash
  & \cellcolor{bestblue}\textbf{10.26} & \cellcolor{bestblue}\textbf{9.88}  & \cellcolor{bestblue}\textbf{10.57}
  & \cellcolor{bestblue}\textbf{7.70}  & \cellcolor{bestblue}\textbf{7.31}  & \cellcolor{bestblue}\textbf{7.94}
  & \cellcolor{bestblue}\textbf{13.69} & \cellcolor{bestblue}\textbf{13.62} & \cellcolor{bestblue}\textbf{12.92}
  & \cellcolor{bestblue}\textbf{30.35} & 40.11 & \cellcolor{bestblue}\textbf{26.91}
  & \cellcolor{bestblue}\textbf{13.74} & 15.10 & \cellcolor{bestblue}\textbf{13.98} \\

Gemini 2.5 Pro
  & 16.53 & 10.66 & 13.93
  & 20.97 & 14.73 & 24.23
  & 25.63 & 19.58 & 24.36
  & 30.44 & 27.85 & 27.33
  & 25.58 & 23.63 & 24.44 \\

Gemma-3-12B-it
  &  47.18 &  95.76 &  47.07
  &  31.50 &  31.81 &  31.67
  & 113.64 &  43.95 &  42.86
  & 111.09 &  65.62 &  60.68
  &  56.69 &  80.51 &  57.25 \\

Gemma-3-4B-it
  &  55.33 &  66.02 &  58.82
  &  52.45 &  46.87 &  44.94
  &  57.48 & 197.39 &  75.11
  & 111.57 &  83.24 &  84.07
  & 174.17 & 128.08 &  76.70 \\

InternVL3.5-2B
  & 111.62 & 172.86 & 112.95
  & 114.79 & 166.29 & 126.73
  & 124.04 & 154.94 & 118.98
  & 137.75 & 238.98 & 134.85
  & 139.02 & 175.51 & 143.29 \\

InternVL3.5-4B
  & 109.66 & 153.91 & 101.33
  & 105.30 & 138.73 & 108.20
  & 118.65 & 169.91 & 112.56
  & 172.73 & 217.80 & 120.47
  & 115.19 & 150.88 & 118.02 \\

InternVL3.5-8B
  & 103.35 & 113.62 &  99.44
  & 107.37 & 108.85 &  98.81
  & 108.32 & 117.44 & 103.83
  & 134.94 & 141.34 & 109.98
  & 118.55 & 115.61 & 104.71 \\

Llama 4 Maverick
  & 26.83 & 24.99 & 27.83
  & 22.85 & 21.71 & 21.79
  & 34.40 & 44.02 & 32.96
  & 56.62 & 51.14 & 60.85
  & 38.61 & 39.51 & 39.69 \\

Qwen3.5-9B
  & 19.59 & 18.35 & 18.21
  & 13.58 & 13.98 & 12.48
  & 17.05 & 16.05 & 16.41
  & 29.29 & 28.50 & 29.73
  & 21.20 & 21.75 & 21.07 \\

Qwen3.5-27B
  & 23.70 & 22.99 & 26.65
  & 15.25 & 14.86 & 15.08
  & 19.14 & 19.31 & 18.85
  & 36.61 & 31.21 & 34.46
  & 24.75 & 24.84 & 24.53 \\

Qwen3.5-35B-A3B
  & 22.61 & 21.08 & 21.46
  & 14.20 & 12.19 & 13.49
  & 17.55 & 15.70 & 16.69
  & 34.19 & \cellcolor{bestblue}\textbf{32.83} & \cellcolor{bestblue}\textbf{32.44}
  & 21.41 & \cellcolor{bestblue}\textbf{20.89} & \cellcolor{bestblue}\textbf{20.83} \\

\bottomrule
\end{tabular}%
}

\smallskip
\noindent\footnotesize
\textbf{P1}: Baseline English \quad
\textbf{P2}: Baseline Bengali \quad
\textbf{P3}: Structured English \quad
Metric: CER (\%, lower is better).
\end{table*}


\definecolor{bestblue}{HTML}{BFDBFE}

\begin{table*}[htbp]
\centering
\caption{%
  Model benchmarking results (CER~\%) on the \textsc{BanglaWild} test set across
  image attribute categories under three prompting strategies:
  \textbf{P1}~(Baseline English), \textbf{P2}~(Baseline Bengali), and
  \textbf{P3}~(Structured English).
  \colorbox{bestblue}{Cyan} highlights the lowest CER per column;
  \textbf{bold} marks the best-performing model per group.%
}
\label{tab:bangla_wild_attributes}
\renewcommand{\arraystretch}{0.88}

\setlength{\tabcolsep}{4.8pt}
\adjustbox{max width=0.95\textwidth}{%
\begin{tabular}{l ccc ccc ccc cc cc cc}
\toprule

& \multicolumn{9}{c}{\textbf{Background Complexity}}
& \multicolumn{6}{c}{\textbf{Curved Text}} \\

\cmidrule(lr){2-10}\cmidrule(lr){11-16}

& \multicolumn{3}{c}{\makecell{\textit{Clean}}}
& \multicolumn{3}{c}{\makecell{\textit{Cluttered}}}
& \multicolumn{3}{c}{\makecell{\textit{Moderate}}}
& \multicolumn{3}{c}{\makecell{\textit{No}}}
& \multicolumn{3}{c}{\makecell{\textit{Yes}}} \\

\cmidrule(lr){2-4}\cmidrule(lr){5-7}\cmidrule(lr){8-10}
\cmidrule(lr){11-13}\cmidrule(lr){14-16}

\textbf{Model}
& P1   & P2   & P3
& P1   & P2   & P3
& P1   & P2   & P3
& P1   & P2   & P3
& P1   & P2   & P3 \\

\midrule

Claude Haiku 4.5
  & 88.17 & 78.48 & 88.02
  & 123.28 & 138.19 & 79.69
  & 93.46 & 85.77 & 96.79
  & 84.79 & 79.11 & 83.34
  & 172.35 & 126.27 & 166.30 \\

Claude Sonnet 4.6
  & 22.11 & 17.10 & \cellcolor{bestblue}\textbf{17.64}
  & 27.79 & 23.03 & \cellcolor{bestblue}\textbf{23.59}
  & 28.32 & 25.82 & 17.96
  & 19.51 & 15.98 & 16.73
  & 75.55 & 54.20 & 35.66 \\

GPT-5.4-mini
  & 51.82 & 54.98 & 50.81
  & 64.65 & 67.58 & 66.19
  & 59.83 & 65.17 & 58.52
  & 51.46 & 53.99 & 50.59
  & 81.52 & 97.87 & 79.81 \\

GPT-5.4-nano
  & 237.46 & 182.33 & 111.78
  & 304.41 & 287.59 & 130.53
  & 288.69 & 227.07 & 129.17
  & 242.06 & 190.91 & 113.94
  & 317.48 & 219.97 & 127.49 \\

Gemini 2.5 Flash
  & \cellcolor{bestblue}\textbf{13.46} & 15.21 & \cellcolor{bestblue}\textbf{13.30}
  & \cellcolor{bestblue}\textbf{24.73} & \cellcolor{bestblue}\textbf{23.80} & 24.98
  & \cellcolor{bestblue}\textbf{14.22} & \cellcolor{bestblue}\textbf{15.13} & \cellcolor{bestblue}\textbf{14.04}
  & \cellcolor{bestblue}\textbf{13.07} & \cellcolor{bestblue}\textbf{14.66} & \cellcolor{bestblue}\textbf{12.90}
  & \cellcolor{bestblue}\textbf{28.69} & \cellcolor{bestblue}\textbf{29.22} & \cellcolor{bestblue}\textbf{28.87} \\

Gemini 2.5 Pro
  & 24.03 & \cellcolor{bestblue}\textbf{20.92} & 23.15
  & 37.05 & 34.76 & 32.97
  & 25.76 & 21.58 & 25.87
  & 24.19 & 20.67 & 23.41
  & 34.46 & 35.77 & 31.66 \\

Gemma-3-12B-it
  & 61.18 & 59.35 & 48.11
  & 64.50 & 75.73 & 86.85
  & 81.78 & 144.94 & 67.67
  & 56.47 & 70.13 & 51.34
  & 170.96 & 79.85 & 66.94 \\

Gemma-3-4B-it
  & 120.13 & 106.48 & 71.11
  & 192.50 & 134.49 & 92.62
  & 211.08 & 201.81 & 73.24
  & 136.25 & 120.63 & 70.88
  & 114.16 & 105.35 & 94.05 \\

InternVL3.5-2B
  & 130.91 & 177.94 & 132.36
  & 138.47 & 162.67 & 171.23
  & 142.69 & 173.64 & 147.52
  & 131.56 & 174.15 & 136.75
  & 149.52 & 213.42 & 126.03 \\

InternVL3.5-4B
  & 117.35 & 155.12 & 113.05
  & 114.57 & 167.61 & 116.15
  & 127.14 & 164.96 & 130.81
  & 117.57 & 154.92 & 115.35
  & 130.93 & 185.80 & 116.15 \\

InternVL3.5-8B
  & 117.25 & 117.33 & 104.91
  & 138.78 & 113.43 & 100.65
  & 102.07 & 115.06 & 99.22
  & 114.88 & 115.64 & 103.04
  & 137.83 & 134.61 & 118.02 \\

Llama 4 Maverick
  & 35.02 & 36.07 & 34.81
  & 58.38 & 65.36 & 57.64
  & 41.85 & 40.44 & 49.82
  & 35.30 & 36.03 & 35.89
  & 60.82 & 66.01 & 64.48 \\

Qwen3.5-9B
  & 19.69 & 19.78 & 19.49
  & 27.94 & 27.94 & 26.83
  & 22.16 & 22.62 & 21.14
  & 18.88 & 19.07 & 18.57
  & 41.98 & 41.29 & 41.27 \\

Qwen3.5-27B
  & 23.35 & 22.44 & 23.15
  & 31.99 & 36.92 & 29.55
  & 24.29 & 25.15 & 25.27
  & 22.23 & 21.77 & 21.99
  & 47.46 & 47.66 & 48.44 \\

Qwen3.5-35B-A3B
  & 20.83 & 19.79 & 20.12
  & 26.52 & 26.12 & 26.12
  & 21.65 & 20.99 & 20.50
  & 19.54 & 18.73 & 18.89
  & 45.06 & 41.87 & 42.89 \\

\bottomrule
\end{tabular}%
}

\medskip

\setlength{\tabcolsep}{5.6pt}
\adjustbox{max width=0.95\textwidth}{%
\begin{tabular}{l ccc ccc ccc ccc ccc ccc}
\toprule

& \multicolumn{9}{c}{\textbf{Font Style}}
& \multicolumn{9}{c}{\textbf{Occlusion}} \\

\cmidrule(lr){2-10}\cmidrule(lr){11-19}

& \multicolumn{3}{c}{\makecell{\textit{Artistic}}}
& \multicolumn{3}{c}{\makecell{\textit{Handwritten}}}
& \multicolumn{3}{c}{\makecell{\textit{Printed}}}
& \multicolumn{3}{c}{\makecell{\textit{Heavy}}}
& \multicolumn{3}{c}{\makecell{\textit{None}}}
& \multicolumn{3}{c}{\makecell{\textit{Partial}}} \\

\cmidrule(lr){2-4}\cmidrule(lr){5-7}\cmidrule(lr){8-10}
\cmidrule(lr){11-13}\cmidrule(lr){14-16}\cmidrule(lr){17-19}

\textbf{Model}
& P1   & P2   & P3
& P1   & P2   & P3
& P1   & P2   & P3
& P1   & P2   & P3
& P1   & P2   & P3
& P1   & P2   & P3 \\

\midrule

Claude Haiku 4.5
  & 82.51 & 66.33 & 86.93
  & 104.05 & 96.98 & 102.58
  & 66.08 & 62.65 & 57.59
  & 258.78 & 211.35 & 78.38
  & 85.97 & 76.21 & 85.88
  & 99.39 & 99.03 & 100.64 \\

Claude Sonnet 4.6
  & 25.57 & 20.93 & 18.14
  & 25.82 & 20.53 & 20.46
  & \cellcolor{bestblue}\textbf{14.40} & \cellcolor{bestblue}\textbf{11.10} & 11.83
  & 23.65 & 24.46 & 22.57
  & 21.46 & 17.36 & 17.24
  & 29.88 & 22.56 & 20.58 \\

GPT-5.4-mini
  & 44.05 & 48.16 & 41.89
  & 62.83 & 67.15 & 62.22
  & 40.30 & 40.91 & 39.84
  & 60.14 & 60.54 & 64.19
  & 51.66 & 54.56 & 50.65
  & 60.06 & 65.75 & 59.21 \\

GPT-5.4-nano
  & 190.65 & 161.74 & 102.22
  & 289.28 & 224.78 & 123.58
  & 202.08 & 147.55 & 106.46
  & 293.11 & 211.89 & 120.27
  & 237.21 & 185.92 & 111.78
  & 283.44 & 219.27 & 126.68 \\

Gemini 2.5 Flash
  & \cellcolor{bestblue}\textbf{14.85} & 18.32 & \cellcolor{bestblue}\textbf{15.03}
  & \cellcolor{bestblue}\textbf{15.17} & 16.70 & \cellcolor{bestblue}\textbf{14.95}
  & 10.73 & \cellcolor{bestblue}\textbf{10.31} & 10.43
  & \cellcolor{bestblue}\textbf{21.08} & \cellcolor{bestblue}\textbf{21.08} & 22.84
  & \cellcolor{bestblue}\textbf{12.87} & \cellcolor{bestblue}\textbf{14.40} & \cellcolor{bestblue}\textbf{12.95}
  & \cellcolor{bestblue}\textbf{18.56} & \cellcolor{bestblue}\textbf{20.11} & \cellcolor{bestblue}\textbf{17.40} \\

Gemini 2.5 Pro
  & 28.45 & 27.24 & 26.41
  & 25.39 & 21.96 & 23.92
  & 20.02 & 15.37 & 21.58
  & 47.03 & 42.70 & 28.24
  & 23.56 & 20.99 & 23.06
  & 28.81 & 23.14 & 27.25 \\

Gemma-3-12B-it
  & 64.60 & 73.82 & 64.82
  & 74.20 & 77.01 & 53.18
  & 38.73 & 52.89 & 38.04
  & 63.51 & 63.51 & 61.08
  & 46.81 & 61.11 & 49.95
  & 132.27 & 109.77 & 61.46 \\

Gemma-3-4B-it
  & 179.72 & 114.05 & 84.55
  & 150.59 & 149.67 & 75.27
  & 53.03 & 53.87 & 53.50
  & 74.05 & 68.92 & 75.54
  & 129.52 & 122.34 & 72.23
  & 159.26 & 111.54 & 72.82 \\

InternVL3.5-2B
  & 127.66 & 156.82 & 133.85
  & 143.42 & 189.20 & 143.73
  & 112.35 & 166.65 & 120.00
  & 131.22 & 143.11 & 168.92
  & 131.06 & 176.25 & 134.08
  & 139.50 & 180.29 & 142.17 \\

InternVL3.5-4B
  & 112.69 & 144.17 & 118.51
  & 123.43 & 164.45 & 118.65
  & 112.26 & 151.66 & 104.61
  & 119.32 & 182.57 & 122.03
  & 117.50 & 154.27 & 113.07
  & 122.13 & 166.16 & 124.38 \\

InternVL3.5-8B
  & 111.68 & 109.26 & 105.74
  & 123.99 & 121.03 & 105.06
  & 102.90 & 114.49 & 99.81
  & 94.59 & 107.84 & 102.43
  & 112.57 & 117.41 & 104.40
  & 132.74 & 115.17 & 102.53 \\

Llama 4 Maverick
  & 33.66 & 33.52 & 36.11
  & 42.20 & 42.49 & 42.99
  & 27.77 & 31.66 & 26.89
  & 65.00 & 96.49 & 67.03
  & 33.57 & 35.40 & 34.66
  & 49.00 & 45.12 & 48.51 \\

Qwen3.5-9B
  & 19.12 & 18.43 & 18.25
  & 22.74 & 23.29 & 22.90
  & 16.02 & 15.97 & 15.03
  & 25.68 & 27.57 & 22.16
  & 19.56 & 19.58 & 19.27
  & 23.39 & 23.86 & 23.03 \\

Qwen3.5-27B
  & 22.45 & 20.93 & 22.29
  & 26.69 & 26.69 & 26.28
  & 18.57 & 18.25 & 19.01
  & 28.11 & 27.97 & 26.49
  & 23.04 & 22.23 & 22.86
  & 26.94 & 28.11 & 26.94 \\

Qwen3.5-35B-A3B
  & 19.69 & 18.78 & 18.75
  & 23.33 & 22.70 & 22.66
  & 17.61 & 15.81 & \cellcolor{bestblue}\textbf{16.86}
  & 24.73 & 24.32 & \cellcolor{bestblue}\textbf{23.65}
  & 20.30 & 19.05 & 19.49
  & 24.59 & 24.76 & 24.13 \\

\bottomrule
\end{tabular}%
}

\smallskip
\noindent\footnotesize
\textbf{P1}: Baseline English \quad
\textbf{P2}: Baseline Bengali \quad
\textbf{P3}: Structured English \quad
Metric: CER (\%, lower is better).
\end{table*}

\begin{table*}[t]\small
\centering
\setlength\tabcolsep{4pt}
\renewcommand{\arraystretch}{1.1}
\resizebox{1.8\columnwidth}{!}{
\begin{tabular}{@{}p{1.5cm} p{1.5cm} p{2.2cm} p{2.7cm} p{2.5cm} p{2.8cm}@{}}
\toprule
\textbf{Ground Truth} & \textbf{Corrected Version} & \textbf{Qwen3.5-9B} & \textbf{Qwen3.5-35B-A3B} & \textbf{Claude Sonnet 4.6} & \textbf{Gemini 2.5 Flash} \\
\midrule

\banglafont{অনুমোতিতে} & \banglafont{অনুমতিতে} &
\banglafont{অনুমোদিতে} & \banglafont{অনুমোতিতে} &
\banglafont{অনুমতিতে} & \banglafont{অনুমতিতে} \\
\textit{anu\underline{mō}tite} & \textit{anu\underline{ma}tite} &
\textit{anu\underline{mod}ite} & \textit{anu\underline{mō}tite} &
\textit{anu\underline{ma}tite} & \textit{anu\underline{ma}tite} \\
\midrule

\banglafont{ধানমণ্ডি} & \banglafont{ধানমন্ডি} &
\banglafont{ধানমন্ডি} & \banglafont{ধানমন্ডি} &
\banglafont{ধানমন্ডি} & \banglafont{ধলমাটি} \\
\textit{dhānama\underline{ṇ}ḍi} & \textit{dhānama\underline{n}ḍi} &
\textit{dhānama\underline{n}ḍi} & \textit{dhānama\underline{n}ḍi} &
\textit{dhānama\underline{n}ḍi} & \textit{dhala\underline{māṭi} (halluc.)} \\
\midrule

\banglafont{বলবীর} & \banglafont{বল বীর} &
\banglafont{বলশীর্} & \banglafont{বন বিক্র} &
\foreignlanguage{hindi}{जल जीव} & \banglafont{বলবীর} \\
\textit{bala\underline{}bīr} & \textit{bala\textvisiblespace{}bīr} &
\textit{bala\underline{śīr}} & \textit{bana\underline{}bikra} &
\textit{jala\underline{}jīva (Hindi)} & \textit{balabīr~\checkmark} \\
\midrule

\banglafont{সুবিদা} & \banglafont{সুবিভা} &
\banglafont{সুবিভা} & \banglafont{সুবিদা} &
\banglafont{সুবিভা} & \banglafont{সুবিধা} \\
\textit{subi\underline{d}ā} & \textit{subi\underline{bh}ā} &
\textit{subi\underline{bh}ā} & \textit{subi\underline{d}ā} &
\textit{subi\underline{bh}ā} & \textit{subi\underline{dh}ā} \\
\midrule

\banglafont{পোষ্টার} & \banglafont{পোস্টার} &
\banglafont{পোস্টার} & \banglafont{পোস্টার} &
\banglafont{পোস্টার} & \banglafont{পোস্টার} \\
\textit{pō\underline{ṣ}ṭār} & \textit{pō\underline{s}ṭār} &
\textit{pō\underline{s}ṭār} & \textit{pō\underline{s}ṭār} &
\textit{pō\underline{s}ṭār} & \textit{pō\underline{s}ṭār} \\
\midrule

\banglafont{ম্যাট্রমনি} & \banglafont{ম্যাট্রিমনি} &
\banglafont{ম্যাট্রমনি} & \banglafont{ম্যাট্রমনি} &
\banglafont{ম্যাট্রিমনি} & \banglafont{ম্যাজিকমণি} \\
\textit{myāṭra\underline{}mni} & \textit{myāṭra\underline{i}mni} &
\textit{myāṭra\underline{}mni} & \textit{myāṭra\underline{}mni} &
\textit{myāṭra\underline{i}mni} & \textit{myā\underline{jikamoṇi} (halluc.)} \\
\midrule

\banglafont{হর্ণ} & \banglafont{হর্ন} &
\banglafont{জন} & \banglafont{তর্ক} &
\banglafont{হর্ন} & \banglafont{শুভ নববর্ষ} \\
\textit{har\underline{ṇ}} & \textit{har\underline{n}} &
\textit{jana (halluc.)} & \textit{tarka (halluc.)} &
\textit{har\underline{n}} & \textit{śubha nababarṣa (halluc.)} \\
\midrule

\banglafont{স্বাধিন} & \banglafont{স্বাধীন} &
\banglafont{স্বাধীন} & \banglafont{অসাধারণ} &
\banglafont{স্বাধিন} & \banglafont{স্বাধীন} \\
\textit{sbādh\underline{i}na} & \textit{sbādh\underline{ī}na} &
\textit{sbādh\underline{ī}na} & \textit{asādh\underline{āraṇa} (halluc.)} &
\textit{sbādh\underline{i}na} & \textit{sbādh\underline{ī}na} \\
\midrule

\banglafont{জীত} & \banglafont{জিত} &
\banglafont{জীবিত} & \banglafont{জীব} &
\banglafont{উৎিত} & \banglafont{জীত} \\
\textit{j\underline{ī}ta} & \textit{j\underline{i}ta} &
\textit{j\underline{ī}bita} & \textit{j\underline{ī}ba} &
\textit{ut\underline{}ita (halluc.)} & \textit{j\underline{ī}ta~\checkmark} \\
\midrule

\banglafont{আষাড়েও} & \banglafont{আষাঢ়েও} &
\banglafont{ফারগুনেই} & \banglafont{আসাদেও} &
\banglafont{ফাগুনেই} & \banglafont{ফাগুনই} \\
\textit{āṣā\underline{ṛ}eo} & \textit{āṣā\underline{ḍh}eo} &
\textit{phāragune\underline{i} (halluc.)} & \textit{āsāde\underline{o}} &
\textit{phāgune\underline{i} (halluc.)} & \textit{phāguni (halluc.)} \\

\bottomrule
\end{tabular}}
\caption{Word-level mismatch analysis between ground truth and corrected version, with corresponding model outputs for the same word. Each row shows the Bengali script (top) and ALA-LC romanization (bottom, \textit{italic}); the differing character(s) are \underline{underlined}. Key orthographic distinctions invisible to non-Bengali readers: \textit{ṇ}~(\banglafont{ণ}, retroflex nasal) vs.\ \textit{n}~(\banglafont{ন}, dental nasal); long vowel \textit{ī}~(\banglafont{ী}) vs.\ short \textit{i}~(\banglafont{ি}); long vowel \textit{ā}~(\banglafont{া}) vs.\ implicit short \textit{a}; retroflex sibilant \textit{ṣ}~(\banglafont{ষ}) vs.\ dental \textit{s}~(\banglafont{স}); and retroflex \textit{ṛ}~(\banglafont{ড়}) vs.\ aspirated \textit{ḍh}~(\banglafont{ঢ়}). These substitutions are counted as character errors in CER. A~\checkmark~denotes output matching the ground truth. \textit{(halluc.)}\ marks outputs with no lexical overlap with the source word. Romanization follows ALA-LC 2017~\cite{alalc2017bengali}.}
\label{tab:mismatch}
\end{table*}

\begin{table*}[t]
\centering
\Large
\setlength{\tabcolsep}{2.5pt}
\renewcommand{\arraystretch}{1.15}
\caption{
Per-category OCR performance for \textbf{Gemini-2.5-Flash} and \textbf{\rev{Qwen3.5-9B}} across all 24 (\textbf{Context} $\times$ \textbf{Content}) groups, under the three prompting regimes:
\textbf{P1} (English baseline), \textbf{P2} (Bengali baseline), and \textbf{P3} (English structured).
All metrics are reported as percentages; lower is better (\textcolor{green}{$\downarrow$}).
For each (model, metric) triple, the best of \{P1, P2, P3\} per row is \textbf{bold}.
Rows are sorted by Gemini's P2 CER, easiest at top.
\del{The bottom row reports the macro-average across all 24 groups. Macro-average is computed over the 24 observed (Context, Content) pairs shown;} \rev{Only the 24 observed (Context, Content) pairs are shown;} 6 of the 30 possible combinations contain no images\del{ and are excluded}.
}
\label{tab:per_category_all_prompts}
\resizebox{2\columnwidth}{!}{%
\begin{tabular}{@{}ll ccc ccc ccc !{\vrule width 0.6pt} ccc ccc ccc@{}}
\toprule
\multirow{3}{*}{\textbf{Context}} & \multirow{3}{*}{\textbf{Content}} &
\multicolumn{9}{c!{\vrule width 0.6pt}}{\textbf{Gemini-2.5-Flash}} &
\multicolumn{9}{c}{{\textbf{\rev{Qwen3.5-9B}}}} \\
\cmidrule(lr){3-11}
\cmidrule(l){12-20}
& & \multicolumn{3}{c}{\textbf{CER}\textcolor{green}{$\downarrow$}} & \multicolumn{3}{c}{\textbf{WER}\textcolor{green}{$\downarrow$}} & \multicolumn{3}{c!{\vrule width 0.6pt}}{\textbf{NED}\textcolor{green}{$\downarrow$}}
  & \multicolumn{3}{c}{\textbf{CER}\textcolor{green}{$\downarrow$}} & \multicolumn{3}{c}{\textbf{WER}\textcolor{green}{$\downarrow$}} & \multicolumn{3}{c}{\textbf{NED}\textcolor{green}{$\downarrow$}} \\
\cmidrule(lr){3-5} \cmidrule(lr){6-8} \cmidrule(lr){9-11}
\cmidrule(lr){12-14} \cmidrule(lr){15-17} \cmidrule(l){18-20}
& & P1 & P2 & P3 & P1 & P2 & P3 & P1 & P2 & P3 & P1 & P2 & P3 & P1 & P2 & P3 & P1 & P2 & P3 \\
\midrule
Printed Material & Instructional & \textbf{2.33} & \textbf{2.33} & 2.52 & \textbf{6.91} & \textbf{6.91} & 7.61 & \textbf{2.30} & \textbf{2.30} & 2.47 & \textbf{3.78} & 5.16 & \textbf{3.78} & \textbf{11.80} & \textbf{11.80} & \textbf{11.80} & \textbf{3.61} & 4.53 & \textbf{3.61} \\
Signboard & Instructional & \textbf{5.81} & 6.10 & 5.94 & \textbf{15.67} & 17.83 & 18.17 & \textbf{5.01} & 5.29 & 5.15 & 9.57 & 9.74 & \textbf{9.20} & \textbf{27.05} & 27.39 & \textbf{27.05} & 8.59 & 8.78 & \textbf{8.25} \\
Printed Material & Advertisement & 10.63 & \textbf{9.93} & 10.73 & \textbf{23.62} & 24.89 & 24.79 & 9.92 & \textbf{9.00} & 9.77 & \textbf{15.73} & 23.40 & 15.99 & 36.65 & 44.05 & \textbf{35.52} & 14.99 & \textbf{14.22} & 14.99 \\
Wall & Religious & 10.28 & \textbf{10.23} & 10.94 & \textbf{31.33} & 31.76 & 32.12 & 9.78 & \textbf{9.76} & 10.35 & \textbf{14.39} & 16.53 & 14.73 & \textbf{41.55} & 43.49 & 41.97 & \textbf{13.92} & 14.99 & 14.28 \\
Printed Material & Creative & 16.95 & \textbf{11.91} & 16.95 & 62.66 & \textbf{35.00} & 62.66 & 13.35 & \textbf{8.96} & 13.35 & \textbf{7.43} & 8.35 & \textbf{7.43} & \textbf{39.58} & 47.91 & \textbf{39.58} & \textbf{7.20} & 8.16 & \textbf{7.20} \\
Printed Material & Religious & 14.56 & \textbf{12.29} & 15.26 & 37.01 & \textbf{36.26} & 40.08 & 12.79 & \textbf{10.49} & 13.36 & 22.44 & 21.91 & \textbf{21.36} & 49.43 & 48.06 & \textbf{47.00} & 21.10 & 20.87 & \textbf{19.98} \\
Wall & Advertisement & \textbf{13.48} & 13.66 & 14.24 & \textbf{33.13} & 34.34 & 34.75 & \textbf{13.06} & 13.17 & 13.77 & 13.22 & \textbf{12.67} & 12.76 & 39.51 & \textbf{38.08} & 38.91 & 12.41 & \textbf{12.07} & 12.42 \\
Packaging & Colloquial & \textbf{14.55} & \textbf{14.55} & 16.36 & \textbf{77.78} & \textbf{77.78} & 88.89 & \textbf{14.55} & \textbf{14.55} & 16.07 & \textbf{30.91} & 36.36 & 32.73 & \textbf{77.78} & 88.89 & 88.89 & \textbf{30.91} & 36.36 & 32.73 \\
Packaging & Advertisement & 16.20 & \textbf{14.70} & 16.06 & 33.73 & \textbf{33.18} & 33.62 & 14.29 & \textbf{12.63} & 14.34 & 25.16 & 22.76 & \textbf{22.75} & 52.31 & 51.26 & \textbf{50.02} & 22.99 & \textbf{20.41} & 21.77 \\
Wall & Creative & \textbf{15.11} & 15.41 & 15.43 & \textbf{28.13} & 28.49 & 28.55 & 14.52 & \textbf{14.37} & 14.57 & \textbf{25.41} & 25.49 & 26.45 & \textbf{48.94} & 49.29 & 49.24 & 24.04 & 24.30 & \textbf{23.70} \\
Printed Material & Colloquial & 19.62 & \textbf{16.27} & 19.68 & 37.62 & \textbf{36.15} & 39.09 & 17.08 & \textbf{14.88} & 17.70 & 38.35 & \textbf{34.73} & 35.58 & 61.38 & \textbf{60.28} & 62.85 & 33.36 & \textbf{32.66} & 34.52 \\
Signboard & Political & 15.29 & 16.35 & \textbf{12.18} & 43.39 & 42.32 & \textbf{36.07} & 14.84 & 15.78 & \textbf{11.71} & \textbf{18.26} & 19.33 & 21.61 & 52.54 & \textbf{44.50} & 58.21 & \textbf{17.78} & 18.45 & 21.09 \\
Printed Material & Political & \textbf{16.76} & 17.59 & \textbf{16.76} & \textbf{20.83} & \textbf{20.83} & \textbf{20.83} & \textbf{14.12} & 15.42 & \textbf{14.12} & 24.02 & 24.02 & \textbf{21.61} & \textbf{37.10} & \textbf{37.10} & 41.27 & 23.99 & 21.49 & \textbf{21.34} \\
Wall & Instructional & 18.79 & \textbf{18.01} & 18.26 & \textbf{32.37} & 32.39 & 33.01 & 13.48 & \textbf{12.98} & 13.37 & 27.41 & \textbf{26.26} & 28.39 & 47.35 & 44.84 & \textbf{44.77} & 25.56 & \textbf{23.80} & 25.97 \\
Signboard & Religious & \textbf{16.13} & 18.32 & 16.22 & 37.76 & \textbf{37.25} & 38.12 & 16.10 & 17.63 & \textbf{16.09} & 20.88 & \textbf{20.05} & 22.41 & 46.45 & 46.55 & \textbf{44.01} & \textbf{18.88} & 19.08 & 19.44 \\
Vehicle & Instructional & 19.53 & 21.51 & \textbf{18.89} & 41.67 & 41.67 & \textbf{38.89} & 19.42 & 20.56 & \textbf{18.78} & 18.76 & 17.93 & \textbf{16.53} & 50.83 & \textbf{43.19} & 48.06 & 18.61 & 17.70 & \textbf{16.33} \\
Signboard & Advertisement & 22.53 & 21.81 & \textbf{20.32} & 40.26 & 40.61 & \textbf{39.90} & 17.57 & 16.88 & \textbf{16.73} & 25.86 & 25.20 & \textbf{24.81} & \textbf{50.31} & 50.54 & 52.00 & 22.42 & \textbf{21.45} & 21.94 \\
Wall & Political & 22.61 & 24.03 & \textbf{22.49} & 39.18 & 41.93 & \textbf{39.07} & 20.09 & \textbf{19.90} & 19.97 & 34.30 & 39.40 & \textbf{33.74} & 61.05 & 65.41 & \textbf{60.67} & 31.02 & \textbf{30.40} & 30.94 \\
Signboard & Colloquial & 27.45 & 26.72 & \textbf{26.58} & 51.30 & \textbf{50.56} & \textbf{50.56} & 26.36 & \textbf{25.81} & 26.10 & 33.01 & \textbf{28.14} & 32.84 & 58.46 & \textbf{53.52} & 59.38 & 31.00 & \textbf{26.67} & 32.19 \\
Vehicle & Advertisement & 31.08 & 32.11 & \textbf{29.26} & 60.25 & 59.34 & \textbf{57.44} & 26.46 & 25.81 & \textbf{25.54} & 31.76 & \textbf{30.17} & 33.41 & 65.69 & \textbf{62.35} & 65.23 & 28.11 & \textbf{27.41} & 28.97 \\
Wall & Colloquial & 33.08 & 34.91 & \textbf{30.05} & 52.65 & 54.16 & \textbf{46.42} & 27.09 & \textbf{25.91} & 26.03 & 40.36 & 43.68 & \textbf{37.93} & \textbf{65.48} & 69.00 & 65.58 & 37.68 & \textbf{35.83} & 37.03 \\
Vehicle & Religious & 31.32 & 44.65 & \textbf{23.11} & \textbf{47.56} & 48.46 & 48.68 & 21.74 & 22.93 & \textbf{21.48} & 26.96 & \textbf{24.88} & 26.90 & \textbf{60.82} & 62.86 & 61.73 & 25.87 & \textbf{23.50} & 25.81 \\
Vehicle & Colloquial & \textbf{33.03} & 47.48 & 33.64 & 65.14 & 77.64 & \textbf{62.82} & \textbf{29.55} & 30.76 & 30.74 & 46.11 & 61.64 & \textbf{45.00} & 76.16 & 91.26 & \textbf{73.24} & 39.90 & \textbf{36.08} & 39.69 \\
Vehicle & Creative & 60.38 & 48.96 & \textbf{48.22} & 59.58 & \textbf{48.33} & 50.83 & 29.35 & 27.78 & \textbf{27.54} & 36.64 & \textbf{32.85} & 36.64 & 64.17 & \textbf{61.67} & 64.17 & 35.17 & \textbf{29.88} & 35.17 \\
\bottomrule
\end{tabular}%
}
\end{table*}

\begin{table}[t]
\centering
\small
\begin{tabular}{llrrr}
\toprule
 & & \multicolumn{3}{c}{\textbf{Exact match (\%)}} \\
\cmidrule(lr){3-5}
\textbf{Model} & \textbf{Prompt} & \textbf{Verbatim} & \textbf{Standard} & \textbf{Neither} \\
\midrule
\multirow{3}{*}{Gemini 2.5 Flash}
 & P1 & 5.03 & 24.85 & 70.12 \\
 & P2 & 3.02 & 26.03 & 70.95 \\
 & P3 & 3.38 & 23.40 & 73.22 \\
\addlinespace

\multirow{3}{*}{Gemini 2.5 Pro}
 & P1 & 5.47 & 24.37 & 70.16 \\
 & P2 & 9.11 & 19.48 & 71.41 \\
 & P3 & 8.28 & 16.15 & 75.57 \\
\addlinespace

\multirow{3}{*}{Claude Sonnet 4.6}
 & P1 & 8.00 & 8.04 & 83.96 \\
 & P2 & 4.64 & 11.38 & 83.98 \\
 & P3 & 7.37 & 12.02 & 80.61 \\
\addlinespace

\multirow{3}{*}{Qwen3.5-35B-A3B}
 & P1 & 4.15 & 6.79 & 89.06 \\
 & P2 & 3.11 & 7.92 & 88.97 \\
 & P3 & 4.15 & 8.15 & 87.70 \\
\addlinespace

\multirow{3}{*}{Llama 4 Maverick}
 & P1 & 4.53 & 6.62 & 90.79 \\
 & P2 & 4.68 & 5.52 & 89.65 \\
 & P3 & 2.22 & 4.98 & 91.00 \\
\addlinespace

\multirow{3}{*}{Qwen3.5-9B}
 & P1 & 3.69 & 4.41 & 92.07 \\
 & P2 & 3.63 & 4.44 & 91.83 \\
 & P3 & 4.11 & 5.43 & 90.39 \\
\addlinespace

\multirow{3}{*}{Qwen3.5-27B}
 & P1 & 0.95 & 4.18 & 94.87 \\
 & P2 & 0.95 & 3.63 & 95.42 \\
 & P3 & 0.95 & 3.63 & 95.42 \\
\addlinespace

\multirow{3}{*}{GPT-5.4-mini}
 & P1 & 0.95 & 0.00 & 99.05 \\
 & P2 & 0.95 & 1.90 & 97.15 \\
 & P3 & 0.95 & 3.80 & 95.25 \\
\addlinespace

Gemma-3-12B-it   & P1--P3 & 0.00 & 0.00 & 100.00 \\
Claude Haiku 4.5 & P1--P3 & 0.00 & 0.00 & 100.00 \\
GPT-5.4-nano     & P1--P3 & 0.00 & 0.00 & 100.00 \\
InternVL3.5-8B   & P1--P3 & 0.00 & 0.00 & 100.00 \\
InternVL3.5-4B   & P1--P3 & 0.00 & 0.00 & 100.00 \\
InternVL3.5-2B   & P1--P3 & 0.00 & 0.00 & 100.00 \\
Gemma-3-4B-it    & P1--P3 & 0.00 & 0.00 & 100.00 \\
\bottomrule
\end{tabular}
\caption{Per-prompt strict exact-match classification on the images whose in-image text
deviates from standard spelling. Each output is classified as an exact match to the
verbatim gold transcription, to the orthographically standard form, or to neither. The
seven models that never produce an exact match under any prompt are collapsed to a single
P1-P3 row. Rows sum to 100 up to rounding.}
\label{tab:exact-match-perprompt}
\end{table}

\end{document}